\pdfoutput=1

\documentclass[11pt]{article}

\usepackage{EMNLP2024}

\usepackage{times}
\usepackage{latexsym}

\usepackage[T1]{fontenc}

\usepackage[utf8]{inputenc}

\usepackage{microtype}

\usepackage{inconsolata}

\usepackage{todonotes}
\usepackage{amsmath}
\usepackage{ntheorem}
\usepackage[inline]{enumitem}
\usepackage{tikz}
\usetikzlibrary{calc, arrows}
\usepackage{caption}
\usepackage{subcaption}
\usepackage{xspace}
\usepackage{booktabs}
\usepackage{rst}
\usepackage{float}
\usepackage{graphicx}
\usepackage{multirow}

\usepackage[capitalize,noabbrev]{cleveref}
\crefname{section}{\S}{\S\S}
\crefname{table}{Tab.}{Tab.}
\crefname{figure}{Fig.}{Figs.}
\crefname{algorithm}{Alg.}{}
\crefname{equation}{Eq.}{Eq.}
\crefname{appendix}{App.}{}
\crefname{theorem}{Theorem}{}
\crefname{restatableTheorem}{Theorem}{}
\crefname{prop}{Proposition}{}
\crefname{definition}{Def.}{}
\crefname{cor}{Corollary}{}
\crefname{observation}{Observation}{}
\crefname{assumption}{Assumption}{}
\crefname{hyp}{Hyp.}{Hypotheses}
\crefformat{section}{\S#2#1#3}
\crefname{namedtheorem}{Hyp.}{Hypotheses}

\newcommand{\defeq}{\overset{\text{\tiny def}}{=}}

\newcommand{\str}{\boldsymbol{u}}
\newcommand{\sym}{u}
\newcommand{\plm}{p}
\newcommand{\eos}{\textsc{eos}}
\newcommand{\alphabet}{\Sigma}
\newcommand{\kleene}[1]{{#1}^\ast}
\newcommand{\clocal}{\boldsymbol{\ell}}
\newcommand{\cglobal}{\boldsymbol{g}}
\newcommand{\sglobal}{s_g}

\newcommand{\pmiacr}{\textsc{pmi}}
\newcommand{\puni}{p_{\text{uni}}}

\newcommand{\defn}[1]{\textbf{#1}}

\newcommand{\deltamse}{{$\Delta\text{MSE}$}}

\newcommand{\hrefEmail}[2]{\href{mailto:#1 }{#2}}

\newtheorem{hypothesis}{Hypothesis}

\title{
Surprise! Uniform Information Density Isn't the Whole Story:\\Predicting Surprisal Contours in Long-form Discourse 
}

\author{Eleftheria Tsipidi \quad Franz Nowak \quad Ryan Cotterell \quad  Ethan Wilcox  \\ 
\textbf{Mario Giulianelli} \quad \textbf{Alex Warstadt} \\ 
ETH Z{\"u}rich \\
\texttt{\{\hrefEmail{tsipidie@ethz.ch}{tsipidie}, \hrefEmail{fnowak@ethz.ch}{fnowak}, \hrefEmail{rcotterell@ethz.ch}{rcotterell}, \hrefEmail{ewilcox@ethz.ch}{ewilcox}, \hrefEmail{mgiulianelli@ethz.ch}{mgiulianelli}, \hrefEmail{warstadt@ethz.ch}{warstadt}\}@ethz.ch}}

\begin{document}
\maketitle
\begin{abstract}
The Uniform Information Density (UID) hypothesis posits that speakers tend to distribute information evenly across linguistic units to achieve efficient communication.
Of course, information rate in texts and discourses is not perfectly uniform.
While these fluctuations can be viewed as theoretically uninteresting noise on top of a uniform target, another explanation is that UID is not the only functional pressure regulating information content in a language.
Speakers may also seek to maintain interest, adhere to writing conventions, and build compelling arguments.
In this paper, we propose one such functional pressure; namely that speakers modulate information rate based on location within a hierarchically-structured model of discourse.
We term this the Structured Context Hypothesis and test it by predicting the surprisal contours of naturally occurring discourses extracted from large language models using predictors derived from discourse structure.
We find that hierarchical predictors are significant predictors of a discourse's information contour and that deeply nested hierarchical predictors are more predictive than shallow ones. 
This work takes an initial step beyond UID to propose testable hypotheses for why the information rate fluctuates in predictable ways.

\vspace{.11em}
\hspace{1.25em}\includegraphics[width=1.25em,height=1.25em]{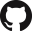}{\hspace{.75em}\parbox{\dimexpr\linewidth-2\fboxsep-2\fboxrule}{\url{https://github.com/rycolab/surprisal-discourse}}}

\end{abstract}

\section{Introduction}\label{sec:introduction}

Linguistic communication takes place in a context, a backdrop of both linguistic and non-linguistic content that can determine how utterances' form \citep{fine2013rapid} and meaning \citep{roberts2006context} are interpreted as well as what words speakers choose to say next \citep{rohde2014grammatical}. 
We investigate the role of context from an information-theoretic perspective, asking how a \defn{linguistic context}, i.e., what has been said or written previously, shapes the information content of each \defn{linguistic unit}, i.e., a novel word or utterance in that context. 
One influential hypothesis for the relationship between linguistic units and their context is the \defn{Uniform Information Density (UID)} hypothesis \citep{fenk1980konstanz,genzel-charniak-2002-entropy,jaeger2006speakers,meister-etal-2021-revisiting, clark-2023-pressure}, which posits that, subject to the constraints of the grammar, speakers spread out information as evenly as possible across an utterance.  
If the UID hypothesis is taken to an extreme, i.e., if it is imposed as a hard constraint, then each linguistic unit would add roughly the same amount of information, when the previous context is taken into account.\looseness=-1

\begin{figure*}
    \centering
    \includegraphics[width=\textwidth] {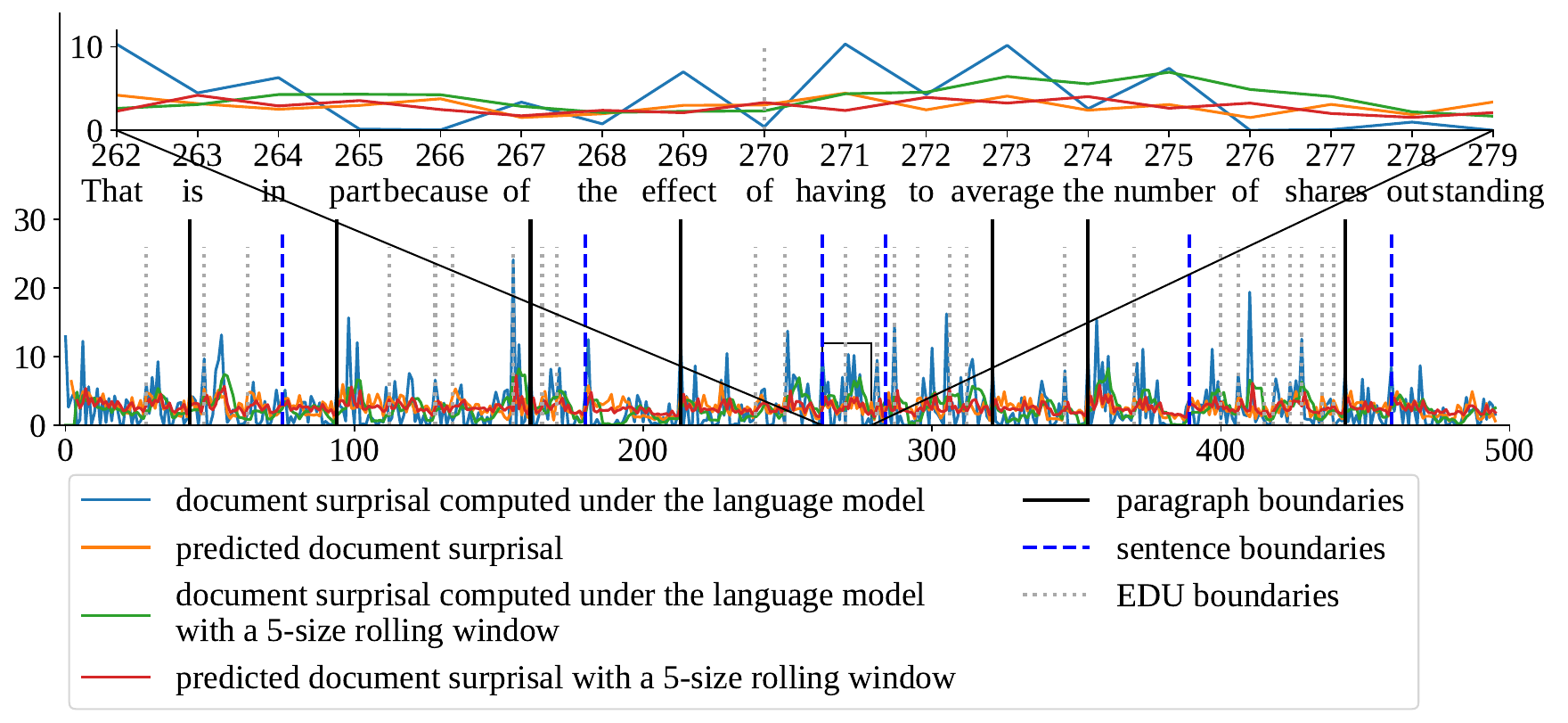}
    \caption{Information contour of the \texttt{wsj\_1111} document from the English RST Discourse Treebank.}
    \label{fig:surprisal_contour}
    \vspace{-8pt}
\end{figure*}

There is an abundance of empirical support for the UID hypothesis, albeit, in general, for a soft variant of it where there is violable \emph{pressure} towards uniformity.
For instance, \citet{clark-2023-pressure} gives evidence across a number of languages that word order is optimized for UID.
Empirically, however, within a discourse, the information content of individual linguistic units is never observed to be strictly static but rather to fluctuate within a band.
We dub this fluctuation in the information content of a discourse its \defn{information contour}; see \Cref{fig:surprisal_contour} for an example.
More theoretically, a pressure towards uniformity must naturally be attenuated by other competing functional pressures on linguistic communication.
Of course, the grammar constrains word choice, which may make uniformity difficult to achieve \citep{jaeger2006speakers}.
Moreover, when an author chooses the next word of a story or a poem, UID might give way to discursive pressures such as a desire for a clean narrative arc or a well-executed rhetorical structure.
Indeed, some literary devices, such as rhyme and meter, may even ascribe higher aesthetic value to a non-uniform information rate.\looseness=-1

In this article, we propose an elaboration of the UID hypothesis. 
In addition to a local pressure for uniformity on information modulated by the grammar, we posit that the information contour of a \emph{discourse} itself is a meaningful signal that reflects a richer structured notion of context.
The idea that there is a relationship between local information content and hierarchical syntax goes back to \citet{hale-2001-probabilistic} and has been expanded more recently \citep{jaffe-etal-2020-coreference,oh2022comparison}.
However, decades of previous research have also established that much like sentences are comprised of syntactical constituents, discourses are organized into nested units as well \citep{mann-thompson-rst-1988,asher2003logics,prasad2008penn}.
Thus, we hypothesize that, in addition to UID, there is a functional pressure on information contours that respects the hierarchical structure of discourse.
We term this the \defn{Structured Context Hypothesis}.
In the context of this hypothesis, we put forth the following research questions:
\begin{enumerate}[label=(\roman*)]
    \item Do structured representations of discourse help explain information contours better than non-structured ones?
    \item And, if so, what type of structure is best at predicting information rates?
\end{enumerate}\looseness=-1

To answer these questions, we use neural language models to estimate the local information content of written English and Spanish texts.
We then consider two different representations for the hierarchical discourse structure of a text.
The first is the standard prose-writing convention of dividing the document into (shallow) hierarchically nested paragraphs and sentences. The second is based on \defn{Rhetorical Structure Theory} (RST; \citealp{mann-thompson-rst-1988}), which breaks texts into recursively nested spans that are linked by discourse relations. 
To investigate questions (i) and (ii) above, we apply Bayesian regression analysis to determine whether access to the discourse structure helps us better model information contours. 
We do find evidence that hierarchical discourse structure helps predict information contours across the board and that RST is more predictive than the shallowly nested paragraph and sentence structures.
In sum, this work provides preliminary empirical evidence for the Structured Context Hypothesis and paves the way for a theory explaining how and why information contours may be modulated by discourse structure.\looseness=-1

\section{Background}
There are myriad information-theoretic theories of language. 
This section builds up to and introduces the Structured Context Hypothesis while contextualizing it in light of previous proposals.\looseness=-1 

\subsection{Language Models and Surprisal}
A \defn{language model} is a probability distribution over $\kleene{\alphabet}$ for a given alphabet $\alphabet$.
Every language model can be decomposed as a product of conditional next-unit probabilities given the units so far, giving an \defn{autoregressive} language model.
Specifically, for any string $\str = \sym_1 \cdots \sym_T \in \kleene{\alphabet}$, we may write 
\begin{equation}
\plm(\str)\defeq\plm(\eos\mid\str)\prod_{t=1}^{T}\plm(\sym_t\mid\str_{<t}).
\end{equation}
Here, $\eos\notin\alphabet$ is a special end-of-string symbol. 

From an autoregressive factorization of a language model, we can define \defn{Shannon surprisal}.
Given a language model $\plm$, the Shannon surprisal \citep{shannon1948mathematical} of a unit in context is its negative log probability in context, i.e., 
\begin{equation}
\label{eq:surprisal}
s(\sym_t) \defeq - \log p(\sym_t \mid \str_{<t}),
\end{equation}
Shannon surprisal (or surprisal for short) is one way to operationalize the notion of a unit's information content under the language model $\plm$, though other operationalizations are possible \citep{giulianelli-etal-2023-information,giulianelli-etal-2024-generalized,giulianelli-etal-2024-incremental}. 
Shannon surprisal as defined above has been hypothesized to correlate with the difficulty of a human reader or listener to process an utterance, a notion known as \defn{surprisal theory} \citep{levy2008expectation} which frames the theory of \citet{hale-2001-probabilistic} in an information-theoretic context.
Specifically, surprisal theory states that the surprisal of a unit quantifies the cost of incrementally updating expectations as a result of observing the unit \citep{levy2008expectation}.
The crucial insight of surprisal theory is that it proposes that, insofar as the language model used to measure the probability of units is a good approximation of the human language model, two distinct properties---information content and processing difficulty---can be quantified with a single metric.\looseness-1

\subsection{Uniform Information Density}
Linguistic communication can be idealized as the transmission of a linguistic signal through a noisy channel with limited capacity \citep[Part 2]{shannon1948mathematical}.
Under this view, a speaker is encouraged to choose a string of linguistic units that contains the most information while not surpassing the channel's capacity.
This functional pressure is one motivation of the UID hypothesis \citep{fenk1980konstanz,jaeger2006speakers}.
The reason for this even distribution is as follows:
On the one hand, if a speaker's linguistic signal contains more information on average than the channel capacity, the communication would be prone to transmission errors. 
On the other, if the information content, on average, were lower than the channel capacity, then there could be an alternative, more efficient way of formulating the linguistic signal. 
The optimal strategy is thus to send information across the channel that is as close to the channel capacity as possible without being too difficult for the comprehender to process.\looseness=-1

The UID hypothesis suggests that production choices aim to optimize both the limitations of channel capacity and the need to efficiently convey information. 
This leads to surprisal being distributed as evenly as possible across a speaker's utterance. 
By preventing significant fluctuations in surprisal, speakers avoid surpassing or falling below channel capacity, ensuring that processing difficulty remains relatively stable for the listener.
The UID hypothesis is supported by empirical studies of language production at the level of syllables \citep{bell2003effects,aylett2004smooth,aylett2006language}, lexical items \citep{meister-etal-2021-revisiting,clark-2023-pressure}, syntactic structures \citep{frank2008speaking,jaeger2010redundancy}, and discourse connectives \citep{torabi-asr-demberg-2012-implicitness,torabi-asr-demberg-2015-uniform}.\looseness=-1

\subsection{Contextualizing the UID Hypothesis}

Most instantiations of the UID hypothesis use the surprisal of a linguistic unit in context as an operationalization of that unit's information content. 
Despite its empirical success at explaining various linguistic phenomena, the UID hypothesis is limited in several ways, which we detail below.

\paragraph{Empirical shortcomings of UID.}

Empirical estimates of character surprisal within words \citep{elman1990finding}, estimates of word surprisal within sentences \citep{levy2013memory,futrell2020lossy} and estimates of sentence surprisal within discourse structures \citep{genzel-charniak-2003-variation} demonstrated that the rate of surprisal fluctuates in ways that correspond to linguistic structure.
For instance, in the case of character surprisal within words, peaks often correspond to morpheme boundaries \citep{Harris1955FromPhoneme,elman1990finding,pimentel-etal-2021-disambiguatory}
and the word surprisal within utterances may correspond to constituent boundaries \cite{jaeger2006speakers}.
However, less work has studied peaks and troughs in information content throughout an entire \emph{discourse}.
We posit that 
the discourse-level fluctuations are likewise not random and may be due to cognitive and linguistic factors.
If information contours fluctuate in a predictable manner, e.g., if they exhibit periodic structure, then this would be evidence against a strong version of the UID hypothesis.\looseness=-1
\paragraph{The Constancy Rate Principle.}
\citet{genzel-charniak-2002-entropy} is one notable example of a study that \emph{does} investigate information contours at the discourse level.
The authors propose the \defn{constancy rate principle}, which stipulates that the \emph{expected} surprisal, i.e., the entropy of the next unit distribution given all previously uttered units, stays roughly constant throughout a discourse.
Specifically, they posit that while the expected surprisal of the next unit given only its current sentence, i.e., taken out of context, increases throughout the discourse, the information contained in the global context grows, too, so that the expected surprisal given the full context stays the same.
As their tools at the time were limited to $n$-gram language models and probabilistic constituency parsers, \citet{genzel-charniak-2002-entropy} could only empirically verify the former claim, i.e., that the surprisal given the local context increases.
More recent studies, however, have exploited Transformer-based models to measure surprisal in the global context.
These studies do find weak evidence of the constancy rate principle, especially when considering languages other than English \citep{verma2023revisiting} or other forms of communication, such as conversation \citep{giulianelli-fernandez-2021-analysing}.
However, even in cases where some constancy is observed, it is always subject to fluctuations within a band that are beyond the explanatory power of the constancy rate principle.\looseness=-1

\paragraph{Other related work.}
Besides uniformity pressures, language production and comprehension are also known to be modulated by discourse structure.
Previous work has investigated how fluctuations of surprisal rates relate to paragraph boundaries \citep{genzel-charniak-2003-variation}, topic shifts in text \citep{qian2011topic} and open-domain dialog \citep{xu-reitter-2016-entropy,maes2022shared}, task-determined contextual units in goal-oriented dialog \citep{giulianelli-etal-2021-information}, as well as extra-linguistic contextual cues \citep{doyle-frank-2015-shared} in multi-party conversations.\looseness=-1

\paragraph{Theoretical shortcomings of UID.}
A second, more theoretical limitation of UID is that it does not inherently take into account language-internal pressures 
other than grammaticality.
Certain linguistic units, regardless of their information profile, might be dispreferred within a linguistic context due to discourse constraints,
argumentative considerations, or aesthetic preferences.
One good example of how language-internal pressures play out at multiple levels of linguistic structure are \defn{contour principles}, constraints against identical segments (or segments with identical features) occurring consecutively which result in non-uniformity of linguistic units. 
Although originally developed to explain non-uniformity of phonological features through the Obligatory Contour Principle \citep{leben1973suprasegmental}, contour principles have been posited to govern the information content of linguistic units at various degrees of granularity, e.g., words within higher levels organization including paragraphs \citep{genzel-charniak-2003-variation} and discourse topics \citep{xu-reitter-2016-entropy}.
In addition, contour-like principles are often recruited to explain, and teach, good writing \citep{kharkwal-muresan-2014-surprisal,snow2015you,archer2016bestseller}. 
At first blush, it is not clear how to reconcile UID with pressures deriving from such contour principles.\looseness=-1

\paragraph{Underspecificity.}
The above discussion points to a broader limitation of UID, namely, that it is underspecified.
While it postulates that information be spread out as evenly as possible throughout linguistic units, it does not provide a specific formulation of uniformity: Which surprisal rates count as uniform? 
And, should information be uniform independently of other language-internal or structural pressures discussed above or only after controlling for these?
Finally, within which notion of linguistic context should surprisal remain uniform? 
Different formulations of uniformity have been explored for rates of word \citep{collins2014information,meister-etal-2021-revisiting} and utterance \citep{giulianelli-fernandez-2021-analysing} surprisal in discourse, with findings hinting at a global uniformity of surprisal---i.e., surprisal tends to stay close to a discourse-level average throughout---especially when larger communicative units are taken into account.

\section{The Structured Context Hypothesis}
To harmonize UID with the constraints imposed by contour principles, we propose the Structured Context Hypothesis.
In most previous work, context is modeled as an essentially sequential object---a succession of paragraphs, topic episodes, dialogue transactions, or dialogue rounds.
In contrast, we rely on a different view, considering context as hierarchical representations made up of sentences within paragraphs or deeply nested discourse trees.
We hypothesize that the fluctuations observed in surprisal contours of discourse beyond a baseline uniformity can be at least partially accounted for by considering structured representations of the discourse in question.
This means that taking into account hierarchical dependencies beyond the sentence level in our theories should increase their ability to predict the information rate of discourse.
We express this view through the following hypothesis:\looseness-1

\begin{hypothesis}
    The Structured Context Hypothesis: The information contour of a discourse is (partially) determined by the hierarchical structure of its constituent discourse units.
\end{hypothesis}
The objective of our experiments is to empirically test this hypothesis on English and Spanish texts.\looseness=-1

In the remainder of this section, we outline two manners to represent documents' hierarchical discourse structure: the conventional prose structure of paragraphs and sentences, and the fine-grained Rhetorical Structure Theory.

\begin{figure}
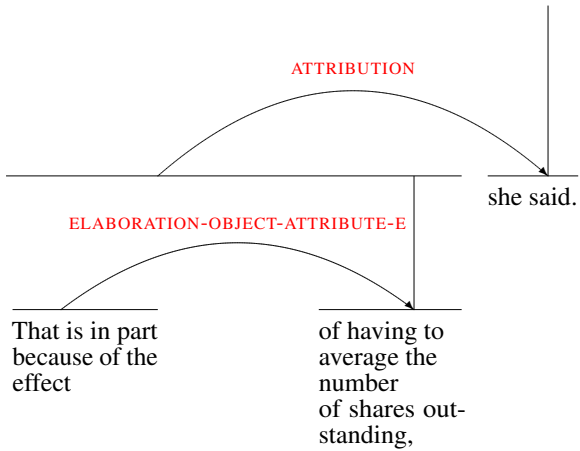

\resizebox{1\columnwidth}{!}{
    \dirrel
{attribution}{\dirrel{elaboration-object-attribute-e}
{\rstsegment{That is in part because of the effect}}
{}{\rstsegment{of having to average the number \\of shares outstanding,}}}
{}{\rstsegment{she said.}}}
    \caption{Discourse sub-tree for a sentence in \texttt{wsj\_1111} from the English RST Discourse Treebank.}   
    \label{fig:en_discourse_subtree}
    \vspace{-15pt}
\end{figure}

\subsection{Conventional Prose Structure}
The first hierarchical discourse structure we consider is the conventional subdivision of documents into paragraphs in which utterances correspond to sentences and the basic linguistic units are individual words.
In what follows, we refer to this way of hierarchically structuring a text simply as \defn{prose structure}.
Documents structured in this way can be seen as shallow trees with a depth of at most three.\looseness=-1

\subsection{Rhetorical Structure Theory}
Rhetorical Structure Theory \citep[RST;][]{mann-thompson-rst-1988} is a well-known discourse analysis framework that posits a high degree of hierarchical structure in a discourse along with categorizing the relationships between parts of the discourse.
The RST representation is a tree structure (\Cref{fig:en_discourse_subtree}); the leaves of the tree correspond to text fragments, usually clauses, which are referred to as \defn{elementary discourse units} (EDUs). 
Internal nodes of the tree correspond to contiguous spans of non-elementary discourse units called \defn{complex discourse units} (CDUs).
While we are primarily concerned with the tree's hierarchical structure, the tree also contains additional information about the text, which may be valuable.
A tree node is labeled as a \defn{nucleus} if it provides essential information, and as a \defn{satellite} if its meaning has a more auxiliary function. 
Tree nodes are also labeled by their \defn{rhetorical relations} to one or more contiguous discourse units, with labels such as \textsc{consequence} or \textsc{elaboration}.\looseness=-1 

\section{Methods}
The goal of our statistical analyses is to study whether the information contour of a text can be predicted from discourse representations.
Our models predict measures of information content (dependent variables) based on a number of predictors (independent variables), some of which we designate as \defn{baseline predictors} while the others designate as \defn{independent predictors} for convenience.
For a summary of all variables, see \Cref{tab:features} in \Cref{app:features-table}.
\looseness=-1
\begin{figure*}[ht!]
\centering
\begin{subfigure}[b]{0.31\textwidth}
\begin{tikzpicture}[->,>=stealth',every label node/.style={text width=1cm}, every circle node/.style={draw, font=\tiny},level 2/.style={sibling distance=38mm},level 3/.style={sibling distance=20mm},level 4/.style={sibling distance=20mm}, every label/.style={font=\scriptsize}, scale=0.66
]
\node (nR){}
child { node[circle, label={[xshift=-1.3cm, yshift=-0.5cm]\tiny {\color{Orange} pop S}, \color{RoyalBlue} push X, Y}] (nS){S}
   child { node[circle, label={[xshift=-1.15cm, yshift=-0.55cm]\tiny {\color{Orange} pop X}, {\color{RoyalBlue} push A, B}}] (nX) {X}
                child { 
                    node[circle, label={[yshift=-1.2cm]({\color{RoyalBlue}5}, {\color{Orange}3})}] (nA) {A}
                    edge from parent node[below left=0.7]{\tiny {\color{Orange} pop A}}
                }
                child { 
                    node[circle, label={[yshift=-1.2cm]({\color{RoyalBlue}5}, {\color{Orange}4})}] (nB) {B}
                    edge from parent node[xshift=0.15cm, below=0.7, left]{\tiny {\color{Orange} pop B}}
                    } 
                edge from parent node[above left=0.2, text width=1.1cm]{} 
            } 
   child { node[circle, label={[xshift=-1.1cm, yshift=-0.55cm]\tiny {{\color{Orange} pop Y}, \color{RoyalBlue} push C, D}}] (nY) {Y}
                child { 
                    node[circle, label={[yshift=-1.2cm]({\color{RoyalBlue}7}, {\color{Orange}6})}] (nC) {C}
                    edge from parent node[xshift=-0.05cm, below left=0.65]{\tiny{\color{Orange} pop C}}
                }
                child {
                    node[circle, label={[yshift=-1.2cm]({\color{RoyalBlue}7}, {\color{Orange}7})}] (nD) {D}
                    edge from parent node[xshift=0.6cm, below left=0.65]{\tiny {\color{Orange} pop D}}
                } 
             } 
    edge from parent node[left]{\tiny {\color{RoyalBlue} push S}}
};

  \draw[->,Gray,rounded corners,dashed,line width=0.7pt]
    ($(nS) + (-0.25,0.4)$) --
    ($(nX) +(-0.35,0.4)$) --
    ($(nX) +(-0.55,0.0)$) --
    ($(nA)  +(-0.35,0.4)$) --
    ($(nA)  +(-0.55,0.1)$) --
    ($(nA)  +(-0.38,-0.38)$) --
    ($(nA)  +(0.0,-0.55)$) --
    ($(nA)  +(0.4,-0.35)$) --
    ($(nA)  +(0.55,0.0)$) --
    ($(nX)  +(0.0,-0.5)$) --
    ($(nB)  +(-0.55,0.1)$) --
    ($(nB)  +(-0.38,-0.38)$) --
    ($(nB)  +(0.0,-0.52)$) --
    ($(nB)  +(0.38,-0.35)$) --
    ($(nB)  +(0.52,0.0)$) --
    ($(nB)  +(0.35,0.35)$) --
    ($(nX)  +(0.4,0.0)$) --
    ($(nS)  +(0.0,-0.4)$) --
    ($(nY)  +(-0.4,0.0)$) --
    ($(nC)  +(-0.35,0.38)$) --
    ($(nC)  +(-0.54,0.1)$) --
    ($(nC)  +(-0.38,-0.38)$) --
    ($(nC)  +(0.0,-0.55)$) --
    ($(nC)  +(0.4,-0.35)$) --
    ($(nC)  +(0.55,0.0)$) --
    ($(nY)  +(0.0,-0.5)$) --
    ($(nD)  +(-0.55,0.1)$) --
    ($(nD)  +(-0.38,-0.38)$) --
    ($(nD)  +(0.0,-0.55)$) --
    ($(nD)  +(0.4,-0.35)$) --
    ($(nD)  +(0.55,0.0)$) --
    ($(nD)  +(0.35,0.35)$) --
    ($(nY) +(0.55,0.0)$) --
    ($(nY) +(0.4,0.35)$) --
    ($(nS) + (0.2,0.52)$);
\end{tikzpicture}
    \caption{top-down parsing}
    \label{fig:rst_pushes_pops_td}
\end{subfigure}
\hfill
\begin{subfigure}[b]{0.31\textwidth}
    \begin{tikzpicture}[->,>=stealth',every label node/.style={text width=1cm}, every circle node/.style={draw, font=\tiny},level 2/.style={sibling distance=35mm},level 3/.style={sibling distance=18mm},level 4/.style={sibling distance=20mm}, every label/.style={font=\scriptsize}, scale=0.66
]
\node (nR){}
child { node[circle, label={[xshift=-0.15cm, yshift=-0.5cm]\tiny{\color{RoyalBlue} push X, Y} \quad\quad\quad\quad\quad\quad{\color{Orange} pop S}}] (nS){S}
   child { node[circle, label={[xshift=-0.17cm, yshift=-0.55cm]\tiny{\color{RoyalBlue} push A, B}\quad\quad\quad\quad{\color{Orange} pop X}}] (nX) {X}
                child { 
                    node[circle, label={[yshift=-1.2cm]({\color{RoyalBlue}5}, {\color{Orange}1})}] (nA) {A}
                    edge from parent node[below left=0.7]{\tiny {\color{Orange} pop A}}
                }
                child { 
                    node[circle, label={[yshift=-1.2cm]({\color{RoyalBlue}5}, {\color{Orange}2})}] (nB) {B}
                    edge from parent node[xshift=0.15cm, below=0.7, left]{\tiny {\color{Orange} pop B}}
                    } 
                edge from parent node[above left=0.2, text width=1.1cm]{} 
            } 
   child { node[circle, label={[xshift=-0.12cm, yshift=-0.55cm]\tiny {\color{RoyalBlue} push C, D}\quad\quad\quad\quad{\color{Orange} pop Y}}] (nY) {Y}
                child { 
                    node[circle, label={[yshift=-1.2cm]({\color{RoyalBlue}7}, {\color{Orange}4})}] (nC) {C}
                    edge from parent node[xshift=-0.01cm, below left=0.65]{\tiny{\color{Orange} pop C}}
                }
                child {
                    node[circle, label={[yshift=-1.2cm]({\color{RoyalBlue}7}, {\color{Orange}5})}] (nD) {D}
                    edge from parent node[xshift=0.1cm, below=0.65, left]{\tiny {\color{Orange} pop D}}
                } 
                edge from parent node[below left=0.2]{} 
             } 
    edge from parent node[left]{\tiny {\color{RoyalBlue} push S}}
};

  \draw[->,Gray,rounded corners,dashed,line width=0.7pt]
    ($(nS) + (-0.25,0.4)$) --
    ($(nX) +(-0.35,0.4)$) --
    ($(nX) +(-0.55,0.0)$) --
    ($(nA)  +(-0.35,0.4)$) --
    ($(nA)  +(-0.55,0.1)$) --
    ($(nA)  +(-0.38,-0.38)$) --
    ($(nA)  +(0.0,-0.55)$) --
    ($(nA)  +(0.4,-0.35)$) --
    ($(nA)  +(0.55,0.0)$) --
    ($(nX)  +(0.0,-0.5)$) --
    ($(nB)  +(-0.55,0.1)$) --
    ($(nB)  +(-0.38,-0.38)$) --
    ($(nB)  +(0.0,-0.52)$) --
    ($(nB)  +(0.38,-0.35)$) --
    ($(nB)  +(0.52,0.0)$) --
    ($(nB)  +(0.35,0.35)$) --
    ($(nX)  +(0.4,0.0)$) --
    ($(nS)  +(0.0,-0.4)$) --
    ($(nY)  +(-0.4,0.0)$) --
    ($(nC)  +(-0.35,0.38)$) --
    ($(nC)  +(-0.54,0.1)$) --
    ($(nC)  +(-0.38,-0.38)$) --
    ($(nC)  +(0.0,-0.55)$) --
    ($(nC)  +(0.4,-0.35)$) --
    ($(nC)  +(0.55,0.0)$) --
    ($(nY)  +(0.0,-0.5)$) --
    ($(nD)  +(-0.55,0.1)$) --
    ($(nD)  +(-0.38,-0.38)$) --
    ($(nD)  +(0.0,-0.55)$) --
    ($(nD)  +(0.4,-0.35)$) --
    ($(nD)  +(0.55,0.0)$) --
    ($(nD)  +(0.35,0.35)$) --
    ($(nY) +(0.55,0.0)$) --
    ($(nY) +(0.4,0.35)$) --
    ($(nS) + (0.2,0.52)$);
\end{tikzpicture}
    \caption{bottom-up parsing}
    \label{fig:rst_pushes_pops_bu}
\end{subfigure}
\begin{subfigure}[b]{0.31\textwidth}
    \begin{tikzpicture}[->,>=stealth',every label node/.style={text width=1cm}, every circle node/.style={draw, font=\tiny},level 2/.style={sibling distance=38mm},level 3/.style={sibling distance=20mm},level 4/.style={sibling distance=20mm}, every label/.style={font=\scriptsize}, scale=0.66
]
\node (nR){}
child { node[circle, label={[xshift=-0.05cm, yshift=-1.05cm, text width=3cm]\tiny{\color{RoyalBlue} push X, Y}\\\quad\\ {\color{Orange}\quad\quad\quad\quad\quad\quad  pop S}}] (nS){S}
   child { node[circle, label={[text width=3cm, xshift=0.16cm, yshift=-1.05cm]\tiny{\color{RoyalBlue} push A, B}\\\quad\\ {\color{Orange}\quad\quad\quad\quad\quad pop X}}] (nX) {X}
                child { 
                    node[circle, label={[yshift=-1.2cm]({\color{RoyalBlue}5}, {\color{Orange}1})}] (nA) {A}
                    edge from parent node[below left=0.7]{\tiny {\color{Orange} pop A}}
                }
                child { 
                    node[circle, label={[yshift=-1.2cm]({\color{RoyalBlue}5}, {\color{Orange}3})}] (nB) {B}
                    edge from parent node[xshift=0.15cm, below=0.7, left]{\tiny {\color{Orange} pop B}}
                    } 
                edge from parent node[above left=0.2, text width=1.1cm]{} 
            } 
   child { node[circle, label={[text width=3cm, xshift=0.16cm, yshift=-1.05cm]\tiny {\color{RoyalBlue} push C, D}\\\quad\\{\color{Orange}\quad\quad\quad\quad\quad pop Y}}] (nY) {Y}
                child { 
                    node[circle, label={[yshift=-1.2cm]({\color{RoyalBlue}7}, {\color{Orange}5})}] (nC) {C}
                    edge from parent node[xshift=-0.05cm, below left=0.65]{\tiny{\color{Orange} pop C}}
                }
                child {
                    node[circle, label={[yshift=-1.2cm]({\color{RoyalBlue}7}, {\color{Orange}7})}] (nD) {D}
                    edge from parent node[xshift=0.1cm, below=0.65, left]{\tiny {\color{Orange} pop D}}
                } 
                edge from parent node[below left=0.2]{} 
             } 
    edge from parent node[left]{\tiny {\color{RoyalBlue} push S}}
    };

  \draw[->,Gray,rounded corners,dashed,line width=0.7pt]
    ($(nS) + (-0.25,0.4)$) --
    ($(nX) +(-0.35,0.4)$) --
    ($(nX) +(-0.55,0.0)$) --
    ($(nA)  +(-0.35,0.4)$) --
    ($(nA)  +(-0.55,0.1)$) --
    ($(nA)  +(-0.38,-0.38)$) --
    ($(nA)  +(0.0,-0.55)$) --
    ($(nA)  +(0.4,-0.35)$) --
    ($(nA)  +(0.55,0.0)$) --
    ($(nX)  +(0.0,-0.5)$) --
    ($(nB)  +(-0.55,0.1)$) --
    ($(nB)  +(-0.38,-0.38)$) --
    ($(nB)  +(0.0,-0.52)$) --
    ($(nB)  +(0.38,-0.35)$) --
    ($(nB)  +(0.52,0.0)$) --
    ($(nB)  +(0.35,0.35)$) --
    ($(nX)  +(0.4,0.0)$) --
    ($(nS)  +(0.0,-0.4)$) --
    ($(nY)  +(-0.4,0.0)$) --
    ($(nC)  +(-0.35,0.38)$) --
    ($(nC)  +(-0.54,0.1)$) --
    ($(nC)  +(-0.38,-0.38)$) --
    ($(nC)  +(0.0,-0.55)$) --
    ($(nC)  +(0.4,-0.35)$) --
    ($(nC)  +(0.55,0.0)$) --
    ($(nY)  +(0.0,-0.5)$) --
    ($(nD)  +(-0.55,0.1)$) --
    ($(nD)  +(-0.38,-0.38)$) --
    ($(nD)  +(0.0,-0.55)$) --
    ($(nD)  +(0.4,-0.35)$) --
    ($(nD)  +(0.55,0.0)$) --
    ($(nD)  +(0.35,0.35)$) --
    ($(nY) +(0.55,0.0)$) --
    ($(nY) +(0.4,0.35)$) --
    ($(nS) + (0.2,0.52)$);
\end{tikzpicture}
    \caption{left-corner parsing}
    \label{fig:rst_pushes_pops_lc}
\end{subfigure}
\caption{Illustration of the \textsc{push}es and \textsc{pop}s from different parsing strategies, with top-down parsing (left) popping nodes in preorder, and bottom-up parsing (right) popping nodes in postorder. Note that the pushing of rules during the depth-first-search is equal in both cases.}
\label{fig:rst_pushes_pops}
\end{figure*}
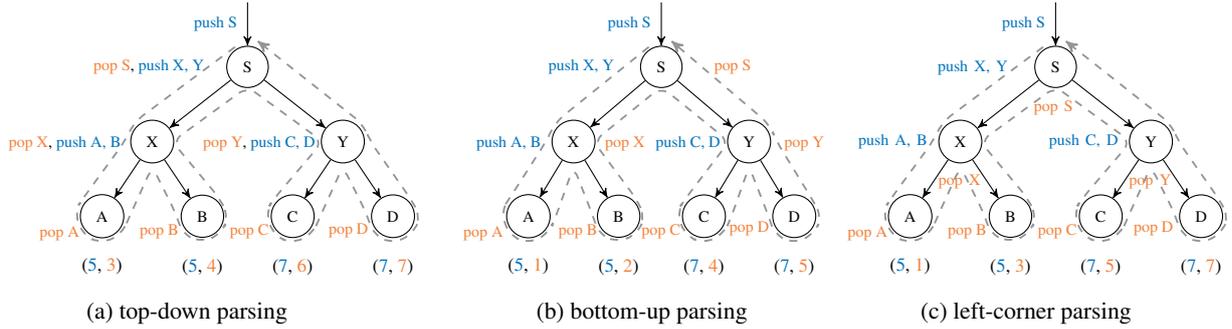

\subsection{Dependent Variables}
\label{sec:method-dependent-variables}
We express information contours in terms of four types of dependent variables (see \Cref{app:dependent-variables} for formal definitions).
The first dependent variable is the global per-unit surprisal, i.e., the surprisal of a unit conditioned on its entire preceding context, starting from the beginning of the document. 
We also refer to this as \defn{document surprisal}. 
In addition to global surprisal, our second dependent variable is its \defn{rolling average} of a window of 3, 5, and 7 units (i.e., tokens).
The third dependent variable type is the difference between a unit's global surprisal and its surprisal in a local context.
This is equivalent to the \defn{pointwise mutual information} (PMI) between the unit and its preceding context conditioned on the local context.
Following previous work \cite[][\textit{inter alia}]{genzel-charniak-2002-entropy,giulianelli-fernandez-2021-analysing}, we consider a local context to be the context beginning with the current sentence or current EDU, and the global context to be all material in a document that precedes the current unit.
We also compute this PMI conditioned on no local context, which is simply the difference of the global surprisal and the unigram surprisal of the unit.
We include these measures to assess how much the particular details of the larger discourse context impact the information content of the current unit.\looseness=-1

\subsection{Baseline Predictors}\label{sec:method-baseline-features}
Our baseline predictors include the length of the current unit, measured in characters, and the surprisal of the previous unit in all experiments.
These are quantities that we expect to be predictive of the current unit's surprisal, but that do not bear directly on the structured context hypothesis.

\subsection{Independent Predictors}\label{sec:method-target-features}
Beyond the baseline predictors, our analyses are based on two main sets of independent predictors: those derived from prose structure trees and those derived from RST trees.
Independent predictors for both types of discourse trees are of four main types.\looseness=-1
\paragraph{Relative position.} These predictors encode the distance of our most granular-level unit\footnote{At the most granular level, our units are tokens, obtained by running the tokenizers of the language models we use to estimate our ground truth surprisal values.} from the beginning of a higher-level structural unit, normalized by the higher-level unit's length; for example, the distance between a token and the start of a paragraph in which it is located, normalized by the number of tokens in the paragraph.\looseness=-1
\paragraph{Nearest boundary.} These predictors encode the distance of a granular-level unit from the closest boundary---left or right---of a higher-level structural unit, normalized by the higher-level unit's length.
Nearest boundary predictors allow us to test for non-monotonic relationships between surprisal and a unit's position in its parent and ancestor units.\looseness-1
\paragraph{Hierarchical position.} These predictors encode the relative position of a unit within its parent in the hierarchical structure, such as the relative position of the unit in a sentence, or of a paragraph (that contains the unit) in the document. 
These predictors allow us to assess the level of a hierarchical context structure that most affects surprisal values.\looseness=-1
\paragraph{Transition predictors.} These predictors encode parsing information on RST and prose structure trees. 
We define integer-valued predictors from the discourse trees yielded by the RST and prose structure annotations of our data. 
We do this by traversing binarized versions of the various trees using common parsing strategies (top-down, bottom-up, and left-corner) for context-free grammars and recording corresponding \textsc{push}
and \textsc{pop} actions between the leaves of the trees; we illustrate this in \Cref{fig:rst_pushes_pops}.
For more details, see \Cref{app:transition}.\looseness=-1

\subsection{Predictive Modeling Framework}\label{sec:predictive-modeling}
To assess the predictive power of different discourse representations, we compare the goodness of fit of a Bayesian linear regressor \citep{clyde2022bayesian} that includes independent and baseline predictors (the \defn{target model}) to one that uses only the baseline predictors (the \defn{baseline model}) to predict information contours. 
Dependent variables and their predictors are described above in \cref{sec:method-dependent-variables,sec:method-baseline-features,sec:method-target-features}; \Cref{app:features-table} provides a summary. 
For each set of predictors, we perform 5-fold cross-validation, estimating a posterior on four folds of the data at a time. 
We fit the Bayesian linear regressor using the using the Pyro framework \citep{bingham2019pyro} with its implementation of stochastic variational inference \citep{hoffman2013stochastic}. 
We use an AutoNormal autoguide,
the Adam optimizer \citep{kingma2014adam}, a learning rate of 0.03, and the evidence lower bound  \citep{kingma2013auto} as our objective.
Then, we compute the expected mean-squared error (MSE) under the Bayesian posterior on the held-out fold.
We aggregate the expected MSEs across the held-out folds to approximate the expected MSE across the entire dataset.
The predictive power of a set of predictors is calculated as the difference in expected MSE between the baseline model and the target model. We refer to this metric as \deltamse. 
To assess the statistical significance of a predictor group's \deltamse, we run paired permutation tests with the cross-validation results.\looseness-1

\section{Data}
We conduct experiments on the English RST Discourse Treebank \citep{carlson-etal-2001-building,carlson2001discourse} and the Spanish RST Discourse Treebank \citep{da-cunha-etal-2011-development}.
For the English Treebank, we consider only the train set, containing 347 documents from the Wall Street Journal.
The Spanish Treebank contains 267 specialist-authored documents in 9 domains, e.g., astrophysics, mathematics, and law; we discard 11 documents due to missing nodes in the RST trees.
\paragraph{Data preprocessing and RST annotations.}
We preprocess the data following \citet{braud-etal-2017-cross}, e.g., we skip document titles which are not part of the RST trees themselves. 
We also use their code\footnote{\href{https://bitbucket.org/chloebt/discourse}{https://bitbucket.org/chloebt/discourse}} to perform right-binarizarization of the RST trees, but do not perform label harmonization \citep[\S4.2]{braud-etal-2017-cross} because we do not make use of any rhetorical relation labels in our experiments.\looseness=-1

\paragraph{Prose structure annotations.}
To recover prose structure boundaries, i.e., paragraph and sentence boundaries, we match English documents to the corresponding plaintexts provided in the Penn Treebank \citep{marcus1999treebank}. 
The Spanish Discourse Treebank directly provides paragraph boundaries, and we recover sentence structure with a text-to-sentence splitter,\footnote{\href{https://github.com/mediacloud/sentence-splitter}{https://github.com/mediacloud/sentence-splitter}} with manual corrections where necessary.
We also perform right-binarization using the NLTK
library \citep{bird2009natural}
to make the prose structure trees consistent with the RST trees.\looseness=-1

\paragraph{Surprisal estimation.}
On the English RST Discourse Treebank, we compute the next-unit surprisal with the \textsc{NousResearch/Yarn-Llama-2-7b-64k} language model \citep{peng2024yarn}.
We selected \textsc{NousResearch/Yarn-Llama-2-7b-64k} because it is trained with a long context window while still being lightweight enough to run on our compute budget.
We compute surprisals on the Spanish RST Discourse Treebanks with the LINCE Mistral 7B Instruct language model.\footnote{\href{https://huggingface.co/NousResearch/Yarn-Llama-2-7b-64k}{https://huggingface.co/NousResearch/Yarn-Llama-2-7b-64k}; \href{https://huggingface.co/clibrain/lince-mistral-7b-it-es}{https://huggingface.co/clibrain/lince-mistral-7b-it-es}}

\begin{figure*}
  \centering
  \begin{subfigure}[b]{0.95\textwidth}
    \includegraphics[width=\textwidth]{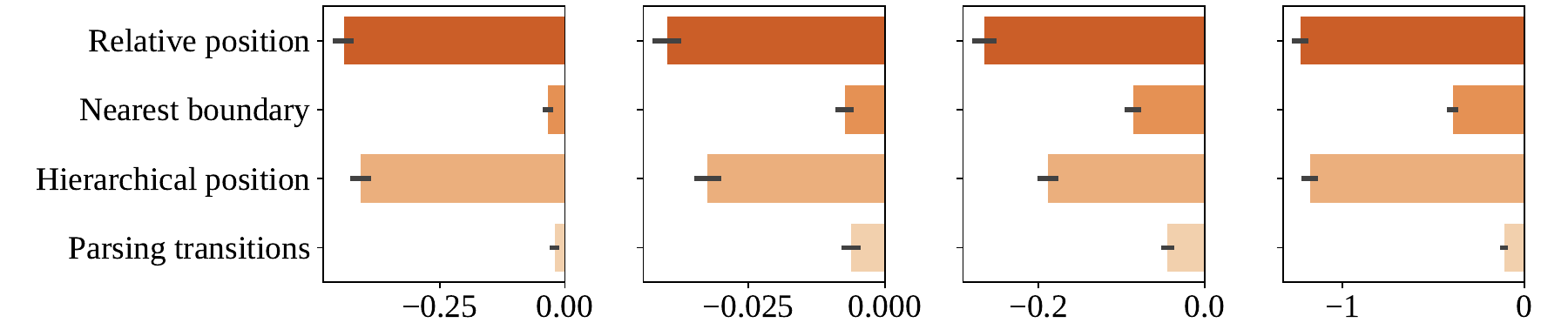}
    \caption{English}
    \label{subfig:rst_comparison_en}
  \end{subfigure}
  \begin{subfigure}[b]{0.95\textwidth}
    \includegraphics[width=\textwidth]{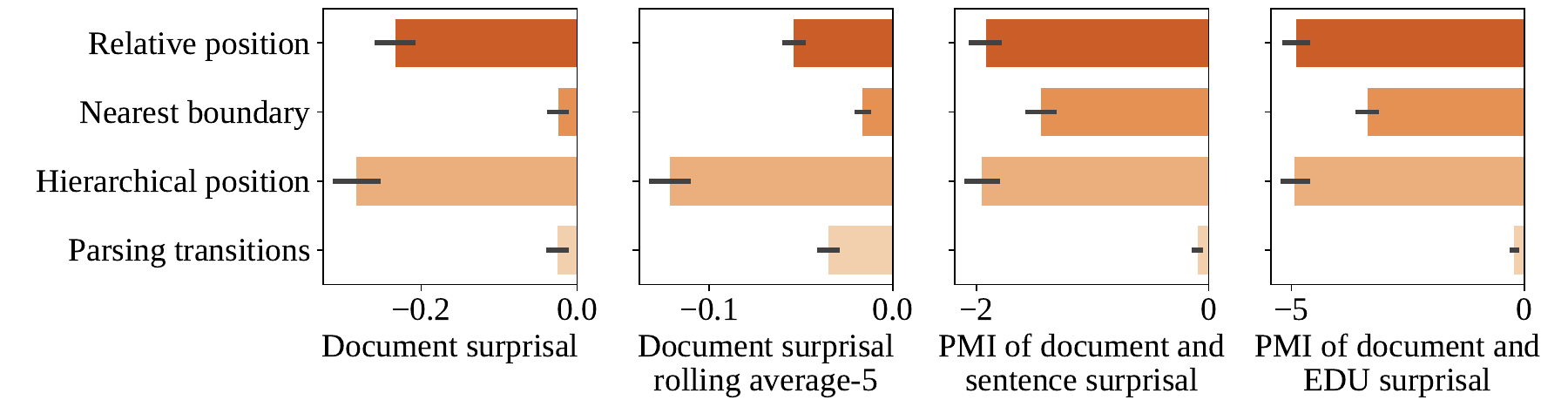}
    \caption{Spanish}
    \label{subfig:rst_comparison_es}
  \end{subfigure}
  \vspace{-8pt}
  \caption{\deltamse\ comparison of models trained on four RST-based predictor groups. Note that the scale for surprisal with a rolling window of 5 is smaller, as rolling average dependent variables exhibit less variance. All these results are statistically significant against the baseline ($p<0.001$).\looseness=-1}
  \label{fig:rst_comparison}
 \vspace{-13pt}
\end{figure*}
\section{Empirical Findings}\label{sec:results}
We overview our empirical results in this section, structuring our presentation in terms of five research questions relating to the Structured Context Hypothesis
In Q1--4, we move from shallower to deeper discourse structure representations, focusing on RST-based predictors.
In Q5, we compare RST to conventional prose structure.

\paragraph{Q1: Are information contours predictable from the relative position within a discourse unit?} \label{par:results-q1}
To answer this question, in \Cref{fig:rst_comparison} we visualize the \deltamse\ (\Cref{sec:predictive-modeling}) of models trained on RST relative position information. We find that including these RST predictors into the model leads to lower \deltamse\ on the held-out data compaared to the baseline ($p<0.001$) indicating that structured contexts help to predict the information contours of a text. Relative position is the best-performing RST-based predictor group for English across dependent variables (\Cref{subfig:rst_comparison_en}; $p<0.001$ against all other predictor groups) and second best for Spanish (\Cref{subfig:rst_comparison_es}; $p<0.001$ against all but hierarchical position).\looseness=-1

\paragraph{Q2: Is the effect of relative position within a discourse unit non-monotonic?} \label{par:results-q2}
To account for possible non-monotonicity, we trained models on predictors including relative distance to nearest boundaries within a discourse unit.
These predictors can account for increases in information content close to the end of the unit after a decrease in the middle of the unit or, vice versa, for lower rates of information content closer to the unit's boundaries.
However, the resulting \deltamse\ for both English and Spanish shows less improvement over the baseline compared to the relative position predictors ($p<0.001$), indicating the effect of a unit's position within a discourse unit is better modeled as monotonic.\looseness=-1
\paragraph{Q3: Does relative position in higher-order structures predict information contours?} \label{par:results-q3}
To assess the explanatory power of hierarchical discourse structures for information contours, we use models that include as predictors the relative position of a unit within its parent in the hierarchical structure.
We find hierarchical position is a significant predictor of all dependent variables analyzed, and either the best or the second-best out of all predictor groups tested.
In the English data, it is moderately less predictive than relative position ($p<0.001$; see  \cref{subfig:rst_comparison_en}).
In the Spanish data, it is the strongest predictor of document surprisal and its rolling average ($p<0.001$) against all other predictors), and on par with the relative position ($p>0.001$) for the PMI dependent variables.\looseness=-1

\paragraph{Q4: Does hierarchical structure encoded by discourse parsing transitions help predict information contours?} \label{par}
As an alternative way to represent the hierarchical structure of the text, we consider predictors obtained by deriving the RST tree structure via constituency parsing algorithms (\Cref{app:transition}).
Although the \deltamse\ is negative in all cases, indicating an increase in predictive power over a baseline model, transition predictors are significantly worse predictors of information contours than relative and hierarchical position ($p<0.001$), for both English and Spanish.\looseness=-1
\begin{figure*}
\centering
\includegraphics[width=\textwidth]{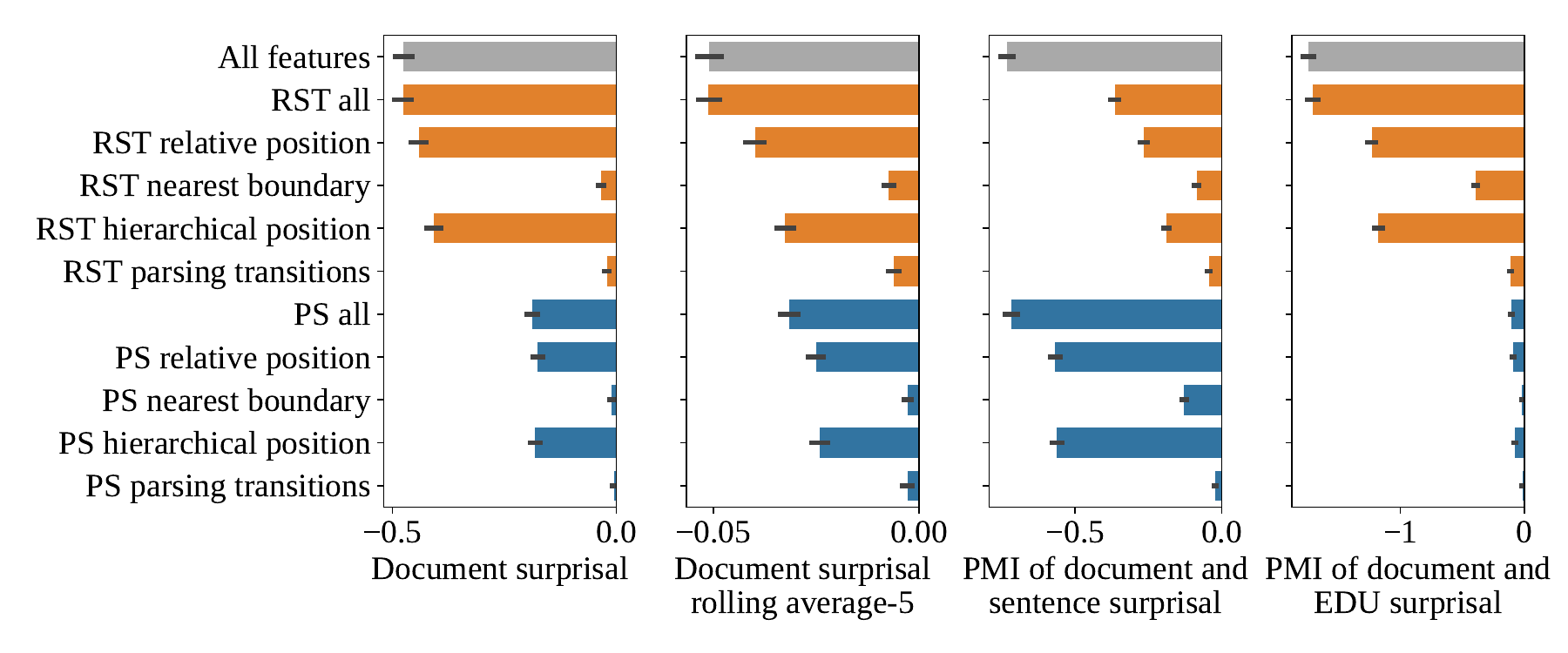}
\vspace{-20pt}
\caption{\deltamse\ across dependent variables of all RST and Prose Structure (PS) predictors on the English data.}
\label{fig:comparison_rst_prose_en}
 \vspace{-5pt}
\end{figure*}
\paragraph{Q5: What representation of discourse structure, RST or Prose Structure, best explains information contours?} \label{par:results-q5}
\Cref{fig:comparison_rst_prose_en} presents a comparison of all individual RST predictors analyzed so far and their prose structure analogs in terms of their \deltamse\ scores across dependent variables on the English data.
We consider two models (referred to as RST all and PS all in \Cref{fig:comparison_rst_prose_en}) that include all predictors derived from either representation of discourse structure.
Results for the Spanish data are shown in \Cref{fig:comparison_rst_prose_es} (\Cref{app:further-experimental-results})\looseness=-1.
Our findings are consistent across the two languages.
For document surprisal and surprisal with a rolling average of 5 units, RST predictors are better than PS predictors ($p<0.001$).
We observe similar trends for PMI of document and unit surprisal, and rolling averages of 3 and 7 (see \Cref{fig:comparison_rst_prose_extra} in \Cref{app:further-experimental-results}).
Furthermore, when considering the locally conditioned PMI variables, we find a correspondence between the strongest family of predictors and the local context over which the PMI is conditioned: The predictive power of RST predictors is higher for EDU-conditioned PMI ($p<0.001$) while PS predictors are better for sentence-conditioned PMI ($p<0.001$).
\paragraph{Summary.} Taken together, our results indicate information contours extracted from language models do exhibit discourse-structural dependencies.
These dependencies are determined both by structural units of conventional prose writing and by more hierarchical discourse units. 
However, explanatory power is higher for the finer-grained and higher-order structures determined by rhetorical relations between discourse units.

\section{Future Work}

We hypothesized that violations of the communicative pressure to communicate at a uniform rate might be predictable, and that part of their predictability is linked to how production choices depend on discourse structure (the Structured Context Hypothesis). 
While we could not determine violations of the UID hypothesis precisely due to its inherent underspecificity, our predictive modeling framework captured deviations from a constant information rate by design, with the intercept representing the baseline rate and predictors capturing deviations. 
As \cref{fig:surprisal_contour} shows, the structured context helps predict oscillations around the base rate, though we only account for a small portion of these deviations.
There are, however, additional intuitive explanations for violations of uniformity which our predictors do not capture, or only do so partially.

\paragraph{Maintaining interest.}
High surprisal content may help to maintain a listener's attention.
In the domain of music synthesis, it has already been proposed that modulating surprisal can affect listener engagement \citep{kothinti2020synthesizing,bjare2024controlling}.
Extending this idea to language, \citet{venkatraman-etal-2023-decoding} found that when controlling for total surprisal, non-uniformity of information density correlates with text quality.

\paragraph{Improving comprehension.}
Overly information-dense content may hinder comprehension.
In such cases, low-surprisal content such as repetitions, reiterations, and summaries at strategic points in the discourse structure can intuitively help to reinforce new information and reduce confusion.
Indeed, redundancy is an important feature in error-correcting codes \citep{hamming1950error}, and repetitions are important for comprehension in noisy-channel situations such as conversations between second-language learners \citep{cervantes1992effects}.

\paragraph{Production constraints.}
Peaks and troughs need not be only out of concern for a listener.
Speakers have limited effort to expend on formulating utterances, and so they may use repetition to maintain the flow of conversation \cite{giulianelli-etal-2022-construction} or hold the floor while formulating a new, more informative utterance \cite{bergey2024producing}.

\paragraph{Aesthetic conventions.}
General aesthetic principles or specific stylistic conventions may intentionally manipulate surprisal. 
Repetition is common to many rhetorical devices \citep{harris2017writing}, and poetic devices such as rhyme and meter increase predictability.
At the level of an entire narrative, emotional arcs have been argued to be conventionalized or to cluster into one of several archetypes \citep{reagan2016emotional,brown2020shapes}, though this idea has not yet, as far as we know, been extended to information content.\\

Each of these explanations draws on intuitions that we believe to be widespread and compelling, yet are out-of-scope for the UID hypothesis. 
More importantly, they make empirical predictions that can be tested by investigating surprisal contours in texts and discourses from different genres that have been annotated for features such as interest and comprehensibility. 
Moving forward, it will become necessary to look at the surprisal contour as just one of many possible types of time-series data that can be associated with a discourse, and which may be related to each other in meaningful ways.\looseness=-1

\section{Conclusion}
We conclude by briefly highlighting the theoretical and empirical contributions of this paper. Theoretically, we have enumerated the limitations of the UID hypothesis and have provided an initial hypothesis, the Structured Context Hypothesis, to predict how information fluctuates during a discourse, namely discourse trees based on prose conventions and RST.
Empirically, we have found support for this hypothesis by evaluating two structured representations of discourse in English and Spanish.
We view this work as one step in developing theories that can explain the vast variation in discourses, texts, and writing genres observed across human cultures.\looseness=-1

\section*{Limitations}
One major limitation of the present work is that it is conducted only in English and Spanish. 
In order to expand to a greater number of languages we have already identified RST-annotated corpora in Basque \citep{iruskieta2013rst,iruskieta2015rst}, Brazilian Portuguese \citep{maziero2015adaptation, cardoso2011cstnews, collovini2007summ,pardo2005rhetalho, pardo2003construccao, pardo2004relaccoes},
Dutch \citep{van2011building, redeker-etal-2012-multi} and German \citep{stede-2004-potsdam,stede-neumann-2014-potsdam,bourgonje-stede-2020-potsdam}. 
These corpora should additionally be tested as a possible next step. 
One other limitation of this work is that we have only used linear models. Although we do investigate whether the relationship between discourse boundaries and surprisal is monotonic, it may be the case that the relationship is non-linear.
Finally, while our theoretical discussion of (non-)uniformity applies to linguistic units of any size, in practice we only measure and predict the surprisal of tokens under the language model (roughly words). 
Our conclusions might change if the surprisals of characters, phonemes, sentences, intonation phrases, or any number of other units are considered.
\looseness=-1

\section*{Ethics Statement}
We foresee no ethical problems with our work.

\bibliography{anthology,custom}

\begin{thebibliography}{89}
\expandafter\ifx\csname natexlab\endcsname\relax\def\natexlab#1{#1}\fi

\bibitem[{Archer and Jockers(2016)}]{archer2016bestseller}
Jodie Archer and Matthew~L. Jockers. 2016.
\newblock \href {https://books.google.com/books?hl=en&lr=&id=F4JzCwAAQBAJ}
  {\emph{The Bestseller Code: {Anatomy} of the Blockbuster Novel}}.
\newblock St. Martin's Press, Inc., USA.

\bibitem[{Asher and Lascarides(2003)}]{asher2003logics}
Nicholas Asher and Alex Lascarides. 2003.
\newblock \href
  {https://www.amazon.com/Conversation-Studies-Natural-Language-Processing/dp/0521659515}
  {\emph{Logics of Conversation}}.
\newblock Cambridge University Press.

\bibitem[{Aylett and Turk(2004)}]{aylett2004smooth}
Matthew Aylett and Alice Turk. 2004.
\newblock \href
  {https://journals.sagepub.com/doi/abs/10.1177/00238309040470010201} {The
  smooth signal redundancy hypothesis: {A} functional explanation for
  relationships between redundancy, prosodic prominence, and duration in
  spontaneous speech}.
\newblock \emph{Language and Speech}, 47(1):31--56.

\bibitem[{Aylett and Turk(2006)}]{aylett2006language}
Matthew Aylett and Alice Turk. 2006.
\newblock \href
  {https://pubs.aip.org/asa/jasa/article-abstract/119/5/3048/893320/Language-redundancy-predicts-syllabic-duration-and}
  {Language redundancy predicts syllabic duration and the spectral
  characteristics of vocalic syllable nuclei}.
\newblock \emph{The Journal of the Acoustical Society of America},
  119(5):3048--3058.

\bibitem[{Bell et~al.(2003)Bell, Jurafsky, Fosler-Lussier, Girand, Gregory, and
  Gildea}]{bell2003effects}
Alan Bell, Daniel Jurafsky, Eric Fosler-Lussier, Cynthia Girand, Michelle
  Gregory, and Daniel Gildea. 2003.
\newblock \href
  {https://pubs.aip.org/asa/jasa/article-abstract/113/2/1001/545131} {Effects
  of disfluencies, predictability, and utterance position on word form
  variation in {E}nglish conversation}.
\newblock \emph{The Journal of the Acoustical Society of America},
  113(2):1001--1024.

\bibitem[{Bergey and DeDeo(2024)}]{bergey2024producing}
Claire~Augusta Bergey and Simon DeDeo. 2024.
\newblock \href {http://arxiv.org/abs/2403.08890} {From "um" to "yeah":
  Producing, predicting, and regulating information flow in human
  conversation}.

\bibitem[{Bingham et~al.(2019)Bingham, Chen, Jankowiak, Obermeyer, Pradhan,
  Karaletsos, Singh, Szerlip, Horsfall, and Goodman}]{bingham2019pyro}
Eli Bingham, Jonathan~P. Chen, Martin Jankowiak, Fritz Obermeyer, Neeraj
  Pradhan, Theofanis Karaletsos, Rohit Singh, Paul~A. Szerlip, Paul Horsfall,
  and Noah~D. Goodman. 2019.
\newblock \href {http://jmlr.org/papers/v20/18-403.html} {Pyro: {Deep}
  universal probabilistic programming}.
\newblock \emph{J. Mach. Learn. Res.}, 20:28:1--28:6.

\bibitem[{Bird et~al.(2009)Bird, Klein, and Loper}]{bird2009natural}
Steven Bird, Ewan Klein, and Edward Loper. 2009.
\newblock \href {https://www.nltk.org/book/} {\emph{Natural Language Processing
  with {P}ython}}.
\newblock O'Reilly Media.

\bibitem[{Bjare et~al.(2024)Bjare, Lattner, and Widmer}]{bjare2024controlling}
Mathias~Rose Bjare, Stefan Lattner, and Gerhard Widmer. 2024.
\newblock \href {http://arxiv.org/abs/2408.06022} {Controlling surprisal in
  music generation via information content curve matching}.
\newblock ArXiv:2408.06022 [cs, eess].

\bibitem[{Bourgonje and Stede(2020)}]{bourgonje-stede-2020-potsdam}
Peter Bourgonje and Manfred Stede. 2020.
\newblock \href {https://aclanthology.org/2020.lrec-1.133} {The {P}otsdam
  commentary corpus 2.2: {Extending} annotations for shallow discourse
  parsing}.
\newblock In \emph{Proceedings of the Twelfth Language Resources and Evaluation
  Conference}, pages 1061--1066, Marseille, France. European Language Resources
  Association.

\bibitem[{Braud et~al.(2017)Braud, Coavoux, and
  S{\o}gaard}]{braud-etal-2017-cross}
Chlo{\'e} Braud, Maximin Coavoux, and Anders S{\o}gaard. 2017.
\newblock \href {https://aclanthology.org/E17-1028} {Cross-lingual {RST}
  discourse parsing}.
\newblock In \emph{Proceedings of the 15th Conference of the {E}uropean Chapter
  of the Association for Computational Linguistics: Volume 1, Long Papers},
  pages 292--304, Valencia, Spain. Association for Computational Linguistics.

\bibitem[{Brown and Tu(2020)}]{brown2020shapes}
Steven Brown and Carmen Tu. 2020.
\newblock \href {https://doi.org/10.1515/fns-2020-0016} {The shapes of stories:
  {A} ``resonator'' model of plot structure}.
\newblock \emph{Frontiers of Narrative Studies}, 6(2):259--288.

\bibitem[{Cardoso et~al.(2011)Cardoso, Maziero, Jorge, Seno, Di~Felippo, Rino,
  Nunes, and Pardo}]{cardoso2011cstnews}
Paula C.~F. Cardoso, Erick~G. Maziero, Mara Luca~Castro Jorge, Eloize M.~R.
  Seno, Ariani Di~Felippo, Lucia Helena~Machado Rino, Maria das Gracas~Volpe
  Nunes, and Thiago A.~S. Pardo. 2011.
\newblock \href
  {http://www.nilc.icmc.usp.br/nilc/download/ariani/CardosoETAL_RST_2011.pdf}
  {{CSTnews-a} discourse-annotated corpus for single and multi-document
  summarization of news texts in {B}razilian {P}ortuguese}.
\newblock In \emph{Proceedings of the 3rd RST Brazilian Meeting}, pages
  88--105.

\bibitem[{Carlson and Marcu(2001)}]{carlson2001discourse}
Lynn Carlson and Daniel Marcu. 2001.
\newblock \href
  {https://web.archive.org/web/20170808131213id_/https://www.isi.edu/~marcu/discourse/tagging-ref-manual.pdf}
  {Discourse tagging reference manual}.
\newblock Technical Report ISI-TR-545, USC Information Sciences Institute.

\bibitem[{Carlson et~al.(2001)Carlson, Marcu, and
  Okurovsky}]{carlson-etal-2001-building}
Lynn Carlson, Daniel Marcu, and Mary~Ellen Okurovsky. 2001.
\newblock \href {https://aclanthology.org/W01-1605} {Building a
  discourse-tagged corpus in the framework of {R}hetorical {S}tructure
  {T}heory}.
\newblock In \emph{Proceedings of the Second {SIG}dial Workshop on Discourse
  and Dialogue}.

\bibitem[{Cervantes and Gainer(1992)}]{cervantes1992effects}
Raoul Cervantes and Glenn Gainer. 1992.
\newblock \href {http://www.jstor.org/stable/3586886} {The effects of syntactic
  simplification and repetition on listening comprehension}.
\newblock \emph{TESOL Quarterly}, 26(4):767--770.
\newblock Wiley.

\bibitem[{Church and Hanks(1990)}]{church-hanks-1990-word}
Kenneth~Ward Church and Patrick Hanks. 1990.
\newblock \href {https://aclanthology.org/J90-1003} {Word association norms,
  mutual information, and lexicography}.
\newblock \emph{Computational Linguistics}, 16(1):22--29.

\bibitem[{Clark et~al.(2023)Clark, Meister, Pimentel, Hahn, Cotterell, Futrell,
  and Levy}]{clark-2023-pressure}
Thomas~Hikaru Clark, Clara Meister, Tiago Pimentel, Michael Hahn, Ryan
  Cotterell, Richard Futrell, and Roger Levy. 2023.
\newblock \href {https://doi.org/10.1162/tacl_a_00589} {A cross-linguistic
  pressure for {U}niform {I}nformation {D}ensity in word order}.
\newblock \emph{Transactions of the Association for Computational Linguistics},
  11:1048--1065.

\bibitem[{Clyde et~al.(2022)Clyde, Çetinkaya Rundel, Rundel, Banks, Chai, and
  Huang}]{clyde2022bayesian}
Merlise Clyde, Mine Çetinkaya Rundel, Colin Rundel, David Banks, Christine
  Chai, and Lizzy Huang. 2022.
\newblock \href {https://statswithr.github.io/book/} {\emph{An Introduction to
  Bayesian Thinking}}, 1 edition.
\newblock Academic Press.

\bibitem[{Collins(2014)}]{collins2014information}
Michael~Xavier Collins. 2014.
\newblock \href
  {https://idp.springer.com/authorize/casa?redirect_uri=https://link.springer.com/article/10.1007/s10936-013-9273-3}
  {Information density and dependency length as complementary cognitive
  models}.
\newblock \emph{Journal of Psycholinguistic Research}, 43:651--681.

\bibitem[{Collovini et~al.(2007)Collovini, Carbonel, Fuchs, Coelho, Rino, and
  Vieira}]{collovini2007summ}
Sandra Collovini, Thiago~I. Carbonel, Juliana~Thiesen Fuchs, Jorge~C{\'e}sar
  Coelho, L{\'u}cia Rino, and Renata Vieira. 2007.
\newblock \href {http://nilc.icmc.usp.br/til/til2007/arq0165.pdf} {Summ-it:
  {Um} corpus anotado com informa{\c{c}}oes discursivas visandoa
  sumariza{\c{c}}ao autom{\'a}tica}.
\newblock \emph{Proceedings of TIL}.

\bibitem[{da~Cunha et~al.(2011)da~Cunha, Torres-Moreno, and
  Sierra}]{da-cunha-etal-2011-development}
Iria da~Cunha, Juan-Manuel Torres-Moreno, and Gerardo Sierra. 2011.
\newblock \href {https://aclanthology.org/W11-0401} {On the development of the
  {RST} {S}panish treebank}.
\newblock In \emph{Proceedings of the 5th Linguistic Annotation Workshop},
  pages 1--10, Portland, Oregon, USA. Association for Computational
  Linguistics.

\bibitem[{Daume~III and Marcu(2002)}]{daume-iii-marcu-2002-noisy}
Hal Daume~III and Daniel Marcu. 2002.
\newblock \href {https://doi.org/10.3115/1073083.1073159} {A noisy-channel
  model for document compression}.
\newblock In \emph{Proceedings of the 40th Annual Meeting of the Association
  for Computational Linguistics}, pages 449--456, Philadelphia, Pennsylvania,
  USA. Association for Computational Linguistics.

\bibitem[{Doyle and Frank(2015)}]{doyle-frank-2015-shared}
Gabriel Doyle and Michael Frank. 2015.
\newblock \href {https://doi.org/10.3115/v1/N15-1182} {Shared common ground
  influences information density in microblog texts}.
\newblock In \emph{Proceedings of the 2015 Conference of the North {A}merican
  Chapter of the Association for Computational Linguistics: Human Language
  Technologies}, pages 1587--1596, Denver, Colorado. Association for
  Computational Linguistics.

\bibitem[{Elman(1990)}]{elman1990finding}
Jeffrey~L. Elman. 1990.
\newblock \href
  {https://www.sciencedirect.com/science/article/abs/pii/036402139090002E}
  {Finding structure in time}.
\newblock \emph{Cognitive Science}, 14(2):179--211.

\bibitem[{Fano(1961)}]{fano1961transmission}
R.~M. Fano. 1961.
\newblock \href {https://books.google.ch/books?id=2PMbtQEACAAJ}
  {\emph{Transmission of Information: A Statistical Theory of Communication}}.
\newblock MIT Press Classics. MIT Press.

\bibitem[{Fenk and Fenk(1980)}]{fenk1980konstanz}
August Fenk and Gertraud Fenk. 1980.
\newblock \href
  {http://wwwg.uni-klu.ac.at/mk0/personal/bedienst/Kurzzeitgedaechtnis1980.pdf}
  {Konstanz im {K}urzzeitged{\"a}chtnis --{K}onstanz im sprachlichen
  {I}nformationsflu{\ss}?}
\newblock \emph{Zeitschrift f{\"u}r experimentelle und angewandte Psychologie},
  27(3):400--414.

\bibitem[{Fine et~al.(2013)Fine, Jaeger, Farmer, and Qian}]{fine2013rapid}
Alex~B. Fine, T.~Florian Jaeger, Thomas~A. Farmer, and Ting Qian. 2013.
\newblock \href
  {https://journals.plos.org/plosone/article?id=10.1371/journal.pone.0077661}
  {Rapid expectation adaptation during syntactic comprehension}.
\newblock \emph{PLOS One}, 8(10):e77661.

\bibitem[{Frank and Jaeger(2008)}]{frank2008speaking}
Austin~F. Frank and T.~Florian Jaeger. 2008.
\newblock \href {https://escholarship.org/content/qt7d08h6j4/qt7d08h6j4.pdf}
  {Speaking rationally: {U}niform information density as an optimal strategy
  for language production}.
\newblock In \emph{Proceedings of the Annual Meeting of the Cognitive Science
  Society}.

\bibitem[{Futrell et~al.(2020)Futrell, Gibson, and Levy}]{futrell2020lossy}
Richard Futrell, Edward Gibson, and Roger~P. Levy. 2020.
\newblock \href {https://doi.org/10.1111/cogs.12814} {Lossy-context surprisal:
  {An} information-theoretic model of memory effects in sentence processing}.
\newblock \emph{Cognitive Science}, 44(3):e12814.

\bibitem[{Genzel and Charniak(2002)}]{genzel-charniak-2002-entropy}
Dmitriy Genzel and Eugene Charniak. 2002.
\newblock \href {https://doi.org/10.3115/1073083.1073117} {Entropy rate
  constancy in text}.
\newblock In \emph{Proceedings of the 40th Annual Meeting of the Association
  for Computational Linguistics}, pages 199--206, Philadelphia, Pennsylvania,
  USA. Association for Computational Linguistics.

\bibitem[{Genzel and Charniak(2003)}]{genzel-charniak-2003-variation}
Dmitriy Genzel and Eugene Charniak. 2003.
\newblock \href {https://aclanthology.org/W03-1009} {Variation of entropy and
  parse trees of sentences as a function of the sentence number}.
\newblock In \emph{Proceedings of the 2003 Conference on Empirical Methods in
  Natural Language Processing}, pages 65--72.

\bibitem[{Gerdemann(1994)}]{gerdemann-1994-parsing}
Dale Gerdemann. 1994.
\newblock \href {https://aclanthology.org/C94-1064} {Parsing as tree
  traversal}.
\newblock In \emph{The 15th International Conference on Computational
  Linguistics}, Kyoto, Japan.

\bibitem[{Giulianelli and
  Fern{\'a}ndez(2021)}]{giulianelli-fernandez-2021-analysing}
Mario Giulianelli and Raquel Fern{\'a}ndez. 2021.
\newblock \href {https://doi.org/10.18653/v1/2021.conll-1.50} {Analysing human
  strategies of information transmission as a function of discourse context}.
\newblock In \emph{Proceedings of the 25th Conference on Computational Natural
  Language Learning}, pages 647--660, Online. Association for Computational
  Linguistics.

\bibitem[{Giulianelli et~al.(2024{\natexlab{a}})Giulianelli, Opedal, and
  Cotterell}]{giulianelli-etal-2024-generalized}
Mario Giulianelli, Andreas Opedal, and Ryan Cotterell. 2024{\natexlab{a}}.
\newblock \href {https://arxiv.org/abs/2409.10728} {Generalized measures of
  anticipation and responsivity in online language processing}.
\newblock In \emph{Findings of the Association for Computational Linguistics:
  EMNLP 2024}, Miami, Florida, USA. Association for Computational Linguistics.

\bibitem[{Giulianelli et~al.(2021)Giulianelli, Sinclair, and
  Fern{\'a}ndez}]{giulianelli-etal-2021-information}
Mario Giulianelli, Arabella Sinclair, and Raquel Fern{\'a}ndez. 2021.
\newblock \href {https://doi.org/10.18653/v1/2021.emnlp-main.652} {Is
  information density uniform in task-oriented dialogues?}
\newblock In \emph{Proceedings of the 2021 Conference on Empirical Methods in
  Natural Language Processing}, pages 8271--8283, Online and Punta Cana,
  Dominican Republic. Association for Computational Linguistics.

\bibitem[{Giulianelli et~al.(2022)Giulianelli, Sinclair, and
  Fern{\'a}ndez}]{giulianelli-etal-2022-construction}
Mario Giulianelli, Arabella Sinclair, and Raquel Fern{\'a}ndez. 2022.
\newblock \href {https://aclanthology.org/2022.aacl-main.51} {Construction
  repetition reduces information rate in dialogue}.
\newblock In \emph{Proceedings of the 2nd Conference of the Asia-Pacific
  Chapter of the Association for Computational Linguistics and the 12th
  International Joint Conference on Natural Language Processing (Volume 1: Long
  Papers)}, pages 665--682, Online only. Association for Computational
  Linguistics.

\bibitem[{Giulianelli et~al.(2024{\natexlab{b}})Giulianelli, Wallbridge,
  Cotterell, and Fernández}]{giulianelli-etal-2024-incremental}
Mario Giulianelli, Sarenne Wallbridge, Ryan Cotterell, and Raquel Fernández.
  2024{\natexlab{b}}.
\newblock \href {https://doi.org/10.31234/osf.io/fhp84} {Incremental
  alternative sampling as a lens into the temporal and representational
  resolution of linguistic prediction}.
\newblock PsyArXiv.

\bibitem[{Giulianelli et~al.(2023)Giulianelli, Wallbridge, and
  Fern{\'a}ndez}]{giulianelli-etal-2023-information}
Mario Giulianelli, Sarenne Wallbridge, and Raquel Fern{\'a}ndez. 2023.
\newblock \href {https://doi.org/10.18653/v1/2023.emnlp-main.343} {Information
  value: Measuring utterance predictability as distance from plausible
  alternatives}.
\newblock In \emph{Proceedings of the 2023 Conference on Empirical Methods in
  Natural Language Processing}, pages 5633--5653, Singapore. Association for
  Computational Linguistics.

\bibitem[{Hale(2001)}]{hale-2001-probabilistic}
John Hale. 2001.
\newblock \href {https://aclanthology.org/N01-1021} {A probabilistic {E}arley
  parser as a psycholinguistic model}.
\newblock In \emph{Second Meeting of the North {A}merican Chapter of the
  Association for Computational Linguistics}.

\bibitem[{Hamming(1950)}]{hamming1950error}
R.~W. Hamming. 1950.
\newblock \href {https://doi.org/10.1002/j.1538-7305.1950.tb00463.x} {Error
  detecting and error correcting codes}.
\newblock \emph{The Bell System Technical Journal}, 29(2):147--160.

\bibitem[{Harris(2017)}]{harris2017writing}
Robert~A. Harris. 2017.
\newblock \href {https://doi.org/10.4324/9780203712047} {\emph{Writing With
  Clarity and Style: A Guide to Rhetorical Devices for Contemporary Writers}}.
\newblock Routledge.

\bibitem[{Harris(1955)}]{Harris1955FromPhoneme}
Zellig~S. Harris. 1955.
\newblock \href {http://www.jstor.org/stable/411036} {From phoneme to
  morpheme}.
\newblock \emph{Language}, 31(2):190--222.

\bibitem[{Hoffman et~al.(2013)Hoffman, Blei, Wang, and
  Paisley}]{hoffman2013stochastic}
Matthew~D. Hoffman, David~M. Blei, Chong Wang, and John Paisley. 2013.
\newblock \href {https://jmlr.org/papers/volume14/hoffman13a/hoffman13a.pdf}
  {Stochastic variational inference}.
\newblock \emph{Journal of Machine Learning Research}.

\bibitem[{Hopcroft et~al.(2001)Hopcroft, Motwani, and Ullman}]{hopcroft01}
John~E. Hopcroft, Rajeev Motwani, and Jeffrey~D. Ullman. 2001.
\newblock \href {https://dl.acm.org/doi/10.1145/568438.568455}
  {\emph{Introduction to Automata Theory, Languages, and Computation}}, 3
  edition.
\newblock Pearson.

\bibitem[{Iruskieta et~al.(2013)Iruskieta, Aranzabe, de~Ilarraza, Gonzalez,
  Lersundi, and de~Lacalle}]{iruskieta2013rst}
Mikel Iruskieta, Mar{\i}a~J Aranzabe, Arantza~Diaz de~Ilarraza, Itziar
  Gonzalez, Mikel Lersundi, and Oier~Lopez de~Lacalle. 2013.
\newblock \href
  {https://ixa.ehu.eus/sites/default/files/dokumentuak/3960/2013RST-Basque-TB.pdf}
  {The {RST} {B}asque {T}ree{B}ank: {an} online search interface to check
  rhetorical relations}.
\newblock In \emph{Proceedings of the 4th Workshop RST and Discourse Studies},
  pages 40--49.

\bibitem[{Iruskieta et~al.(2015)Iruskieta, de~Ilarraza, and
  Lersundi}]{iruskieta2015rst}
Mikel Iruskieta, Arantza~Diaz de~Ilarraza, and Mikel Lersundi. 2015.
\newblock \href {https://doi.org/doi:10.1515/cllt-2013-0008} {Establishing
  criteria for {RST-based} discourse segmentation and annotation for texts in
  {B}asque}.
\newblock \emph{Corpus Linguistics and Linguistic Theory}, 11(2):303--334.

\bibitem[{Jaeger and Levy(2006)}]{jaeger2006speakers}
T.~Jaeger and Roger Levy. 2006.
\newblock \href
  {https://papers.nips.cc/paper_files/paper/2006/hash/c6a01432c8138d46ba39957a8250e027-Abstract.html}
  {Speakers optimize information density through syntactic reduction}.
\newblock \emph{Advances in Neural Information Processing Systems}, 19.

\bibitem[{Jaeger(2010)}]{jaeger2010redundancy}
T.~Florian Jaeger. 2010.
\newblock \href
  {https://www.sciencedirect.com/science/article/pii/S0010028510000083}
  {Redundancy and reduction: {S}peakers manage syntactic information density}.
\newblock \emph{Cognitive Psychology}, 61(1):23--62.

\bibitem[{Jaffe et~al.(2020)Jaffe, Shain, and
  Schuler}]{jaffe-etal-2020-coreference}
Evan Jaffe, Cory Shain, and William Schuler. 2020.
\newblock \href {https://doi.org/10.18653/v1/2020.coling-main.404} {Coreference
  information guides human expectations during natural reading}.
\newblock In \emph{Proceedings of the 28th International Conference on
  Computational Linguistics}, pages 4587--4599, Barcelona, Spain (Online).
  International Committee on Computational Linguistics.

\bibitem[{Johnson and Roark(2000)}]{johnson-roark-2000-compact}
Mark Johnson and Brian Roark. 2000.
\newblock \href {https://aclanthology.org/C00-1052} {Compact non-left-recursive
  grammars using the selective left-corner transform and factoring}.
\newblock In \emph{{COLING} 2000 Volume 1: The 18th International Conference on
  Computational Linguistics}.

\bibitem[{Joty et~al.(2012)Joty, Carenini, and Ng}]{joty-etal-2012-novel}
Shafiq Joty, Giuseppe Carenini, and Raymond Ng. 2012.
\newblock \href {https://aclanthology.org/D12-1083} {A novel discriminative
  framework for sentence-level discourse analysis}.
\newblock In \emph{Proceedings of the 2012 Joint Conference on Empirical
  Methods in Natural Language Processing and Computational Natural Language
  Learning}, pages 904--915, Jeju Island, Korea. Association for Computational
  Linguistics.

\bibitem[{Kharkwal and Muresan(2014)}]{kharkwal-muresan-2014-surprisal}
Gaurav Kharkwal and Smaranda Muresan. 2014.
\newblock \href {https://doi.org/10.3115/v1/W14-1807} {Surprisal as a predictor
  of essay quality}.
\newblock In \emph{Proceedings of the Ninth Workshop on Innovative Use of {NLP}
  for Building Educational Applications}, pages 54--60, Baltimore, Maryland.
  Association for Computational Linguistics.

\bibitem[{Kingma and Ba(2015)}]{kingma2014adam}
Diederik~P. Kingma and Jimmy Ba. 2015.
\newblock \href {https://arxiv.org/abs/1412.6980} {Adam: A method for
  stochastic optimization}.
\newblock In \emph{3rd International Conference on Learning Representations},
  San Diego, CA, USA.

\bibitem[{Kingma and Welling(2014)}]{kingma2013auto}
Diederik~P. Kingma and Max Welling. 2014.
\newblock \href {https://arxiv.org/abs/1312.6114} {Auto-encoding variational
  {Bayes}}.
\newblock In \emph{2nd International Conference on Learning Representations}.

\bibitem[{Kothinti et~al.(2020)Kothinti, Skerritt-Davis, Nair, and
  Elhilali}]{kothinti2020synthesizing}
Sandeep Kothinti, Benjamin Skerritt-Davis, Aditya Nair, and Mounya Elhilali.
  2020.
\newblock \href {https://ieeexplore.ieee.org/document/9054500/} {Synthesizing
  engaging music using dynamic models of statistical surprisal}.
\newblock In \emph{International Conference on Acoustics, Speech and Signal
  Processing}, pages 761--765, Barcelona, Spain. IEEE.

\bibitem[{Leben(1973)}]{leben1973suprasegmental}
William~Ronald Leben. 1973.
\newblock \href {http://www.ai.mit.edu/projects/dm/theses/leben73.pdf}
  {\emph{Suprasegmental Phonology}}.
\newblock Ph.D. thesis, Massachusetts Institute of Technology.

\bibitem[{Levy(2008)}]{levy2008expectation}
Roger Levy. 2008.
\newblock \href
  {https://www.sciencedirect.com/science/article/pii/S0010027707001436?casa_token=84k3XCwSwVkAAAAA:u0CbJMhg05o5hF-zk1ktgH375Y8Zo0y7Qn3MUt-3R0Q07zCUUTLUFJ799xxXJarZ1aOdt250tcs}
  {Expectation-based syntactic comprehension}.
\newblock \emph{Cognition}, 106(3):1126--1177.

\bibitem[{Levy(2013)}]{levy2013memory}
Roger Levy. 2013.
\newblock \href
  {https://pages.ucsd.edu/~bkbergen/cogs200/sentence-processing-book-chapter.pdf}
  {Memory and surprisal in human sentence comprehension}.
\newblock In \emph{Sentence Processing}, pages 78--114. Psychology Press.

\bibitem[{Ma{\"e}s et~al.(2022)Ma{\"e}s, Blache, and
  Becerra-Bonache}]{maes2022shared}
Eliot Ma{\"e}s, Philippe Blache, and Leonor Becerra-Bonache. 2022.
\newblock \href {https://hal.science/hal-04151675/document} {Shared knowledge
  in natural conversations: {c}an entropy metrics shed light on information
  transfers?}
\newblock In \emph{26th Conference on Computational Natural Language Learning},
  pages 213--227.

\bibitem[{Mann and Thompson(1988)}]{mann-thompson-rst-1988}
William~C. Mann and Sandra~A. Thompson. 1988.
\newblock \href {https://doi.org/doi:10.1515/text.1.1988.8.3.243} {Rhetorical
  structure theory: {Toward} a functional theory of text organization}.
\newblock \emph{Text - Interdisciplinary Journal for the Study of Discourse},
  8(3):243--281.

\bibitem[{Marcus et~al.(1999)Marcus, Santorini, Marcinkiewicz, and
  Taylor}]{marcus1999treebank}
Mitchell~P. Marcus, Beatrice Santorini, Mary~Ann Marcinkiewicz, and Ann Taylor.
  1999.
\newblock \href {https://catalog.ldc.upenn.edu/LDC99T42} {Treebank-3}.
\newblock \emph{Linguistic Data Consortium}, 14.

\bibitem[{Maziero et~al.(2015)Maziero, Hirst, and
  Pardo}]{maziero2015adaptation}
Erick~G. Maziero, Graeme Hirst, and Thiago~A.S. Pardo. 2015.
\newblock \href {https://doi.org/10.1109/BRACIS.2015.24} {Adaptation of
  discourse parsing models for the {P}ortuguese language}.
\newblock In \emph{2015 Brazilian Conference on Intelligent Systems (BRACIS)},
  pages 140--145.

\bibitem[{Meister et~al.(2021)Meister, Pimentel, Haller, J{\"a}ger, Cotterell,
  and Levy}]{meister-etal-2021-revisiting}
Clara Meister, Tiago Pimentel, Patrick Haller, Lena J{\"a}ger, Ryan Cotterell,
  and Roger Levy. 2021.
\newblock \href {https://doi.org/10.18653/v1/2021.emnlp-main.74} {Revisiting
  the {U}niform {I}nformation {D}ensity hypothesis}.
\newblock In \emph{Proceedings of the 2021 Conference on Empirical Methods in
  Natural Language Processing}, pages 963--980, Online and Punta Cana,
  Dominican Republic. Association for Computational Linguistics.

\bibitem[{Oh et~al.(2022)Oh, Clark, and Schuler}]{oh2022comparison}
Byung-Doh Oh, Christian Clark, and William Schuler. 2022.
\newblock \href
  {https://www.frontiersin.org/journals/artificial-intelligence/articles/10.3389/frai.2022.777963/full}
  {Comparison of structural parsers and neural language models as surprisal
  estimators}.
\newblock \emph{Frontiers in Artificial Intelligence}, 5:777963.

\bibitem[{Opedal et~al.(2024)Opedal, Chodroff, Cotterell, and
  Wilcox}]{opedal2024rolecontextreadingtime}
Andreas Opedal, Eleanor Chodroff, Ryan Cotterell, and Ethan Wilcox. 2024.
\newblock \href {http://arxiv.org/abs/2409.08160} {On the role of context in
  reading time prediction}.
\newblock In \emph{Proceedings of the 2024 Conference on Empirical Methods in
  Natural Language Processing}, Miami, Florida, USA. Association for
  Computational Linguistics.

\bibitem[{Opedal et~al.(2023)Opedal, Tsipidi, Pimentel, Cotterell, and
  Vieira}]{opedal-etal-2023-exploration}
Andreas Opedal, Eleftheria Tsipidi, Tiago Pimentel, Ryan Cotterell, and Tim
  Vieira. 2023.
\newblock \href {https://doi.org/10.18653/v1/2023.emnlp-main.827} {An
  exploration of left-corner transformations}.
\newblock In \emph{Proceedings of the 2023 Conference on Empirical Methods in
  Natural Language Processing}, pages 13393--13427, Singapore. Association for
  Computational Linguistics.

\bibitem[{Pardo and Nunes(2003)}]{pardo2003construccao}
Thiago Alexandre~Salgueiro Pardo and Maria das Gra{\c{c}}as~Volpe Nunes. 2003.
\newblock \href
  {https://repositorio.usp.br/directbitstream/a198382e-6171-43f5-9517-c17ddbc70c9e/BIBLIOTECA_113_RT_212.pdf}
  {A constru{\c{c}}{\~a}o de um corpus de textos cient{\'\i}ficos em
  portugu{\^e}s do brasil e sua marca{\c{c}}{\~a}o ret{\'o}rica}.
\newblock Technical report, Universidade de São Paulo.

\bibitem[{Pardo and Nunes(2004)}]{pardo2004relaccoes}
Thiago Alexandre~Salgueiro Pardo and Maria das Gra{\c{c}}as~Volpe Nunes. 2004.
\newblock \href
  {https://repositorio.usp.br/directbitstream/42f0508b-58f3-4977-a98d-43325df0b402/BIBLIOTECA_113_RT_231.pdf}
  {Rela{\c{c}}{\~o}es ret{\'o}ricas e seus marcadores superficiais::
  an{\'a}lise de um corpus de textos cient{\'\i}ficos em portugu{\^e}s do
  brasil}.
\newblock \emph{Relat{\'o}rio T{\'e}cnico NILC}.

\bibitem[{Pardo and Seno(2005)}]{pardo2005rhetalho}
Thiago Alexandre~Salgueiro Pardo and Eloize Rossi~Marques Seno. 2005.
\newblock \href
  {https://sites.icmc.usp.br/taspardo/VEncontroCorpora2005-PardoSeno.pdf}
  {Rhetalho: {um} corpus de refer{\^e}ncia anotado retoricamente}.
\newblock \emph{Proceedings of Encontro de Corpora}.

\bibitem[{Peng et~al.(2024)Peng, Quesnelle, Fan, and Shippole}]{peng2024yarn}
Bowen Peng, Jeffrey Quesnelle, Honglu Fan, and Enrico Shippole. 2024.
\newblock \href {https://openreview.net/forum?id=wHBfxhZu1u} {Ya{RN}:
  {Efficient} context window extension of large language models}.
\newblock In \emph{The Twelfth International Conference on Learning
  Representations}.

\bibitem[{Pimentel et~al.(2021)Pimentel, Cotterell, and
  Roark}]{pimentel-etal-2021-disambiguatory}
Tiago Pimentel, Ryan Cotterell, and Brian Roark. 2021.
\newblock \href {https://doi.org/10.18653/v1/2021.eacl-main.3} {Disambiguatory
  signals are stronger in word-initial positions}.
\newblock In \emph{Proceedings of the 16th Conference of the European Chapter
  of the Association for Computational Linguistics: Main Volume}, pages 31--41,
  Online. Association for Computational Linguistics.

\bibitem[{Prasad et~al.(2008)Prasad, Dinesh, Lee, Miltsakaki, Robaldo, Joshi,
  and Webber}]{prasad2008penn}
Rashmi Prasad, Nikhil Dinesh, Alan Lee, Eleni Miltsakaki, Livio Robaldo,
  Aravind Joshi, and Bonnie Webber. 2008.
\newblock \href
  {http://www.lrec-conf.org/proceedings/lrec2008/pdf/754_paper.pdf} {The {P}enn
  {D}iscourse {T}ree{B}ank 2.0.}
\newblock In \emph{Proceedings of the Sixth International Conference on
  Language Resources and Evaluation ({LREC}'08)}, Marrakech, Morocco. European
  Language Resources Association (ELRA).

\bibitem[{Qian and Jaeger(2011)}]{qian2011topic}
Ting Qian and T.~Florian Jaeger. 2011.
\newblock \href {https://escholarship.org/content/qt6b0712jg/qt6b0712jg.pdf}
  {Topic shift in efficient discourse production}.
\newblock In \emph{Proceedings of the Annual Meeting of the Cognitive Science
  Society}, volume~33.

\bibitem[{Reagan et~al.(2016)Reagan, Mitchell, Kiley, Danforth, and
  Dodds}]{reagan2016emotional}
Andrew~J. Reagan, Lewis Mitchell, Dilan Kiley, Christopher~M. Danforth, and
  Peter~Sheridan Dodds. 2016.
\newblock \href {https://doi.org/10.1140/epjds/s13688-016-0093-1} {The
  emotional arcs of stories are dominated by six basic shapes}.
\newblock \emph{EPJ Data Science}, 5(31).

\bibitem[{Redeker et~al.(2012)Redeker, Berzl{\'a}novich, van~der Vliet, Bouma,
  and Egg}]{redeker-etal-2012-multi}
Gisela Redeker, Ildik{\'o} Berzl{\'a}novich, Nynke van~der Vliet, Gosse Bouma,
  and Markus Egg. 2012.
\newblock \href
  {http://www.lrec-conf.org/proceedings/lrec2012/pdf/887_Paper.pdf}
  {Multi-layer discourse annotation of a {D}utch text corpus}.
\newblock In \emph{Proceedings of the Eighth International Conference on
  Language Resources and Evaluation}, pages 2820--2825, Istanbul, Turkey.
  European Language Resources Association.

\bibitem[{Roberts(2006)}]{roberts2006context}
Craige Roberts. 2006.
\newblock \href
  {https://onlinelibrary.wiley.com/doi/pdf/10.1002/9780470756959#page=212}
  {Context in dynamic interpretation}.
\newblock \emph{The Handbook of Pragmatics}, pages 197--220.

\bibitem[{Rohde and Kehler(2014)}]{rohde2014grammatical}
Hannah Rohde and Andrew Kehler. 2014.
\newblock \href
  {https://www.tandfonline.com/doi/full/10.1080/01690965.2013.854918}
  {Grammatical and information-structural influences on pronoun production}.
\newblock \emph{Language, Cognition and Neuroscience}, 29(8):912--927.

\bibitem[{Rosenkrantz and Lewis(1970)}]{rosenkrantz1970}
D.~J. Rosenkrantz and P.~M. Lewis. 1970.
\newblock \href {https://doi.org/10.1109/SWAT.1970.5} {Deterministic left
  corner parsing}.
\newblock In \emph{11th Annual Symposium on Switching and Automata Theory (swat
  1970)}, pages 139--152.

\bibitem[{Shannon(1948)}]{shannon1948mathematical}
Claude~Elwood Shannon. 1948.
\newblock \href {https://dl.acm.org/doi/pdf/10.1145/584091.584093} {A
  mathematical theory of communication}.
\newblock \emph{The Bell System Technical Journal}, 27(3):379--423.

\bibitem[{Snow et~al.(2015)Snow, Allen, Jacovina, Perret, and
  McNamara}]{snow2015you}
Erica~L. Snow, Laura~K. Allen, Matthew~E. Jacovina, Cecile~A. Perret, and
  Danielle~S. McNamara. 2015.
\newblock \href {https://doi.org/10.1145/2723576.2723592} {You've got style:
  {Detecting} writing flexibility across time}.
\newblock In \emph{Proceedings of the Fifth International Conference on
  Learning Analytics And Knowledge}, page 194–202, New York, NY, USA.
  Association for Computing Machinery.

\bibitem[{Stede(2004)}]{stede-2004-potsdam}
Manfred Stede. 2004.
\newblock \href {https://aclanthology.org/W04-0213} {The {P}otsdam commentary
  corpus}.
\newblock In \emph{Proceedings of the Workshop on Discourse Annotation}, pages
  96--102, Barcelona, Spain. Association for Computational Linguistics.

\bibitem[{Stede and Neumann(2014)}]{stede-neumann-2014-potsdam}
Manfred Stede and Arne Neumann. 2014.
\newblock \href
  {http://www.lrec-conf.org/proceedings/lrec2014/pdf/579_Paper.pdf} {{P}otsdam
  commentary corpus 2.0: {Annotation} for discourse research}.
\newblock In \emph{Proceedings of the Ninth International Conference on
  Language Resources and Evaluation}, pages 925--929, Reykjavik, Iceland.
  European Language Resources Association (ELRA).

\bibitem[{Torabi~Asr and Demberg(2012)}]{torabi-asr-demberg-2012-implicitness}
Fatemeh Torabi~Asr and Vera Demberg. 2012.
\newblock \href {https://aclanthology.org/C12-1163} {Implicitness of discourse
  relations}.
\newblock In \emph{Proceedings of the 24th International Conference on
  Computational Linguistics}, pages 2669--2684, Mumbai, India.

\bibitem[{Torabi~Asr and Demberg(2015)}]{torabi-asr-demberg-2015-uniform}
Fatemeh Torabi~Asr and Vera Demberg. 2015.
\newblock \href {https://aclanthology.org/W15-0117} {Uniform surprisal at the
  level of discourse relations: {N}egation markers and discourse connective
  omission}.
\newblock In \emph{Proceedings of the 11th International Conference on
  Computational Semantics}, pages 118--128, London, UK. Association for
  Computational Linguistics.

\bibitem[{Van Der~Vliet et~al.(2011)Van Der~Vliet, Berzl{\'a}novich, Bouma,
  Egg, and Redeker}]{van2011building}
Nynke Van Der~Vliet, Ildik{\'o} Berzl{\'a}novich, Gosse Bouma, Markus Egg, and
  Gisela Redeker. 2011.
\newblock \href {https://d-nb.info/1119241413/34#page=165} {Building a
  discourse-annotated {D}utch text corpus}.
\newblock \emph{S. Dipper and H. Zinsmeister (Eds.), Beyond Semantics, Bochumer
  Linguistische Arbeitsbericht}, 3:157--171.

\bibitem[{Venkatraman et~al.(2023)Venkatraman, He, and
  Reitter}]{venkatraman-etal-2023-decoding}
Saranya Venkatraman, He~He, and David Reitter. 2023.
\newblock \href {https://doi.org/10.18653/v1/2023.findings-eacl.70} {How do
  decoding algorithms distribute information in dialogue responses?}
\newblock In \emph{Findings of the Association for Computational Linguistics:
  EACL 2023}, pages 953--962, Dubrovnik, Croatia. Association for Computational
  Linguistics.

\bibitem[{Verma et~al.(2023)Verma, Tomlin, and Klein}]{verma2023revisiting}
Vivek Verma, Nicholas Tomlin, and Dan Klein. 2023.
\newblock \href {https://doi.org/10.18653/v1/2023.findings-emnlp.1039}
  {Revisiting entropy rate constancy in text}.
\newblock In \emph{Findings of the Association for Computational Linguistics:
  EMNLP 2023}, pages 15537--15549, Singapore. Association for Computational
  Linguistics.

\bibitem[{Xu and Reitter(2016)}]{xu-reitter-2016-entropy}
Yang Xu and David Reitter. 2016.
\newblock \href {https://doi.org/10.18653/v1/P16-1051} {Entropy converges
  between dialogue participants: {E}xplanations from an information-theoretic
  perspective}.
\newblock In \emph{Proceedings of the 54th Annual Meeting of the Association
  for Computational Linguistics (Volume 1: Long Papers)}, pages 537--546,
  Berlin, Germany. Association for Computational Linguistics.

\end{thebibliography}
\bibliographystyle{acl_natbib}

\clearpage

\appendix
\begin{table*}[t]
    \centering
    \resizebox{\textwidth}{!}{
    \begin{tabular}{p{0.25\textwidth}p{0.15\textwidth}p{0.55\textwidth}}
    \toprule
        \bf Variable Family & \bf Variable Type & \bf Description \\
    \midrule
        Document surprisal & Dependent & Surprisal of unit $\sym$ with global context $\cglobal$ \\
        Rolling average ($n$) & Dependent & Rolling average of document surprisal with a window $n \in \{3, 5, 7\}$\\
        PMI & Dependent & Pointwise mutual information of:\\
        ~&~&(i) $\sym$ with global context $\cglobal$ and $\sym$ without context (unigram)\\
        ~&~&(ii) $\sym$ with global context $\cglobal$ and $\sym$ with local context $\clocal$ (i.e., the containing sentence in prose structure, or the containing EDU in RST) \\
    \midrule
        Relative position & Independent & Relative position of unit $\sym$ within higher-level unit \\
        Boundary distance & Independent & Relative distance of $\sym$ from the nearest boundary (start or end) of higher level unit \\
        Hierarchical position & Independent & Relative position of discourse unit $v$ (where $v$ is or contains $\sym$) within higher-level unit $w$ normalized by the total number of discourse units nested directly under $w$ \\
        Parsing transitions & Independent & \{previous, next\} $\times$ \{\textsc{push}es, \textsc{pop}s\} $\times$ \{bottom-up, top-down, left corner\} number of transitions of either type directly preceding or following $\sym$ according to different parsing strategies\\
    \midrule
        Unit length & Baseline & length of $\sym$ in terms of lower-level units\\
        Previous unit surprisal & Baseline & Surprisal of unit preceding $\sym$ \\
    \bottomrule
    \end{tabular}%
    }
    \caption{Summary of all the variables (dependent variables, independent predictors, and baseline predictors) used in our regression analysis. All variables are associated with a single unit $\sym$.\looseness=-1}
    \label{tab:features}
\end{table*}

\section{Reproducibility}
We extracted the true surprisal values using an RTX 4090 GPU with VRAM 24GB and additional RAM of 64GB for 6 hours. Our predictive modeling experiments required a total of 70 hours on an RTX 3080 GPU with a 10 GB VRAM and 32GB RAM. For details on autoguides, we refer to Pyro documentation\footnote{\href{https://docs.pyro.ai/en/dev/infer.autoguide.html}{https://docs.pyro.ai/en/dev/infer.autoguide.html}}. 

\section{Overview of Variables}
\label{app:features-table}
\Cref{tab:features} provides an overview of the independent and dependent variables used in our experiments.

\section{Transition Predictors}
\label{app:transition}

\paragraph{RST transition predictors.}

To incorporate the hierarchical structure information of the RST annotations, we extract several integer variables from the RST trees corresponding to their tree structure.
In line with related work \citep[\textit{inter alia}]{daume-iii-marcu-2002-noisy, joty-etal-2012-novel}, we assume that RST annotations correspond to parse trees of a context-free grammar (CFG).
Most CFG constituency parsers are of one of three variants that determine in which order the nodes of the parse tree are constructed: 
Top-down (TD), bottom-up (BU), and left-corner (LC) \citep{rosenkrantz1970, johnson-roark-2000-compact, opedal-etal-2023-exploration}.
As \citet{gerdemann-1994-parsing} notes, each of the three parsing variants follows a specific depth-first-search tree traversal strategy.
Specifically, TD, BU, and LC correspond to pre-order, post-order, and in-order traversal for a given parse tree, respectively.\looseness=-1

We first preprocess the RST trees by right-binarizing them.
Then, for each of the parsing strategies, we assign integer values to the leaves of the RST trees using the following steps:
\begin{enumerate*}[label=\arabic*)]
    \item Traverse the RST tree in depth-first order;
    \item When adding a CFG rule to the set of rules to be evaluated later, we increment a \textsc{push} counter;
    \item When evaluating a rule at a node, we increment a \textsc{pop} counter;
    \item When reaching a leaf node, we assign that node the value of the \textsc{push} and \textsc{pop} counters.
\end{enumerate*}
In other words, each RST terminal node gets assigned a tuple containing the number of \textsc{push}es and \textsc{pop}s that happened before evaluating the leaf node's rule under TD, BU, and LC parsing. 
Note that this is related to how pushdown automata parse context-free grammars \citep[Ch. 6]{hopcroft01}.
Since in bottom-up parsing, not all \textsc{pop}s happen before the last EDU of a document, we report the actions twice for each EDU, recording both the \defn{previous} actions up to the given EDU and the \defn{next} actions which are the same values but shifted by one position to the left.
See \Cref{fig:rst_pushes_pops} for an illustration.\looseness=-1

\paragraph{Prose structure Transition predictors.}
We also extract transition predictors from the prose structure of our data using the same method.
The main difference is that the structural units of prose structure are sentences and paragraphs rather than RST EDUs.
To perform constituency parsing on the (flat) prose structure trees, we first right-binarize them, as we did for the RST trees.
The transition predictors can then be extracted using the same rules as described above since the parsing strategies work for arbitrary binary trees.

\section{Dependent Variables} \label{app:dependent-variables}
The goal of our analyses is to test whether the information rate of text can be predicted from discourse trees.
We express information rate in terms of four types of dependent variables.
We consider a document to be made up of hierarchically arranged units, where each higher-level unit contains the units below it in the structure. 
We use the following notation: $\sym$ is a unit drawn from an alphabet $\alphabet$, and $\str$ is a string of units, i.e., an element of $\alphabet^*$. 
Note that we consider the alphabet $\alphabet$ of units to correspond to a \emph{specific} level of the discourse tree; e.g., characters, words, sentences, etc. 
When looking at such a string of same-level units in a hierarchical document, each individual unit can be contextualized as a tuple $(\sym, \clocal, \cglobal)$, where $\cglobal$ is the \defn{global context}, i.e., all the units that linearly preceded $\sym$ in the document, and $\clocal$ is the \defn{local context}, i.e., all the units that preceded $\sym$ in the document with the additional restriction that they share the same parent in the hierarchical structure.
Note the global context subsumes the local context. 
When we use prose structure to compute the dependent variables, the units are tokens and the parent units are sentences, meaning the global context $\cglobal$ of a token $\sym$ contains all the tokens before $\sym$ in the document, while the local context $\clocal$ contains all the tokens from the start of the sentence.
When we use RST trees, the local context of a unit is all the preceding units in the given EDU.\looseness=-1

\paragraph{Global surprisal.} This is the per-unit surprisal conditioned on the entire preceding context, starting from the beginning of the document:
\begin{equation}\label{eq:global-surprisal}
    \sglobal (\sym) \defeq - \log \plm(\sym \mid \cglobal),
\end{equation}
where $\plm$ is the probability produced by a language model. 
We will also refer to global surprisal as \defn{document surprisal} in experiments.
\Cref{eq:global-surprisal} is identical to \Cref{eq:surprisal}.

\paragraph{Rolling average of global surprisal.} 
We compute the rolling average of document information contours over windows of size $n \in \{3, 5, 7\}$. 
Thus, the highly local peaks and throughs of the original information contour are smoothened out in the resulting contours.\looseness=-1

\paragraph{PMI: Unit and global context.} 
We also measure the difference between a unit's unigram probability and its global surprisal under our language model. 
This difference is the pointwise mutual information \citep[\pmiacr;][]{fano1961transmission} between the unit and its global context:\looseness=-1
\begin{align}
    \pmiacr (\sym ; \cglobal) = \log \puni(\sym) - \log \plm(\sym \mid \cglobal).
\end{align}
where $\puni$ is $\sym$'s unigram probability \citep[Eq. 10b]{opedal2024rolecontextreadingtime}.
$\pmiacr$ is a common measure in NLP \citep{church-hanks-1990-word} and measures the degree of association, or mutual dependence, between the two variables.

\paragraph{PMI: Unit and global context conditioned on local context.}
We also measure the $\pmiacr$ between a unit and its global context, when the local context is taken into account:
\begin{equation}
\begin{aligned}
\pmiacr (\sym &; \cglobal \mid \clocal) \\
&= \log \plm(\sym \mid \clocal) -\log \plm(\sym \mid \cglobal, \clocal).
\end{aligned}
\end{equation}
This is a measure of how much larger discourse context impacts the information of the current unit, even when local information is taken into account.
Specifically, we take units to be tokens and compute two versions of this value, one where the local context is the containing sentence and one where it is the containing EDU.\looseness=-1

\section{Further Experimental Results}\label{app:further-experimental-results}
\begin{figure*}[h]
\centering
 \includegraphics[width=\textwidth]{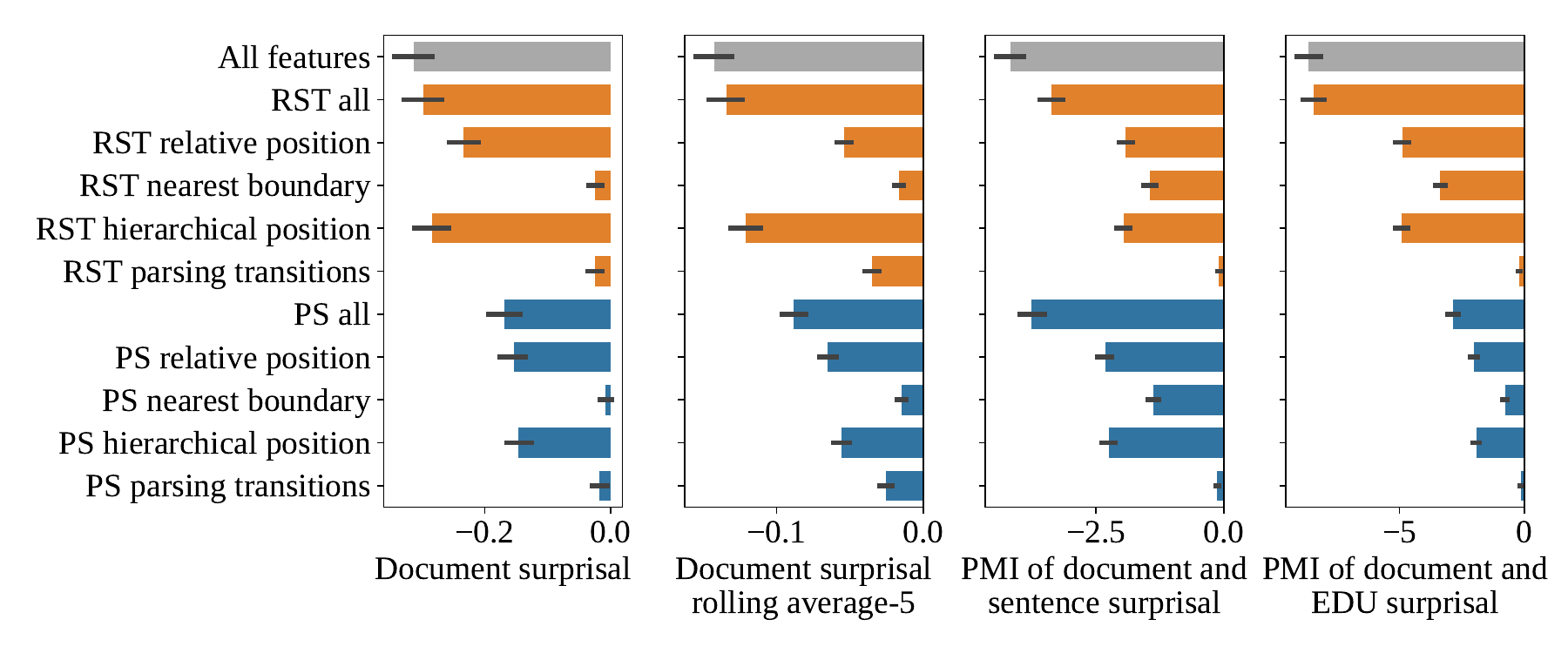}
  \caption{\deltamse\ of RST and Prose Structure (PS) across the same dependent variables as \Cref{fig:comparison_rst_prose_en} on Spanish data.}
  \label{fig:comparison_rst_prose_es}
\end{figure*}

\begin{figure*}[ht!]
\centering
\begin{subfigure}[b]{1\textwidth}
 \includegraphics[width=\textwidth]{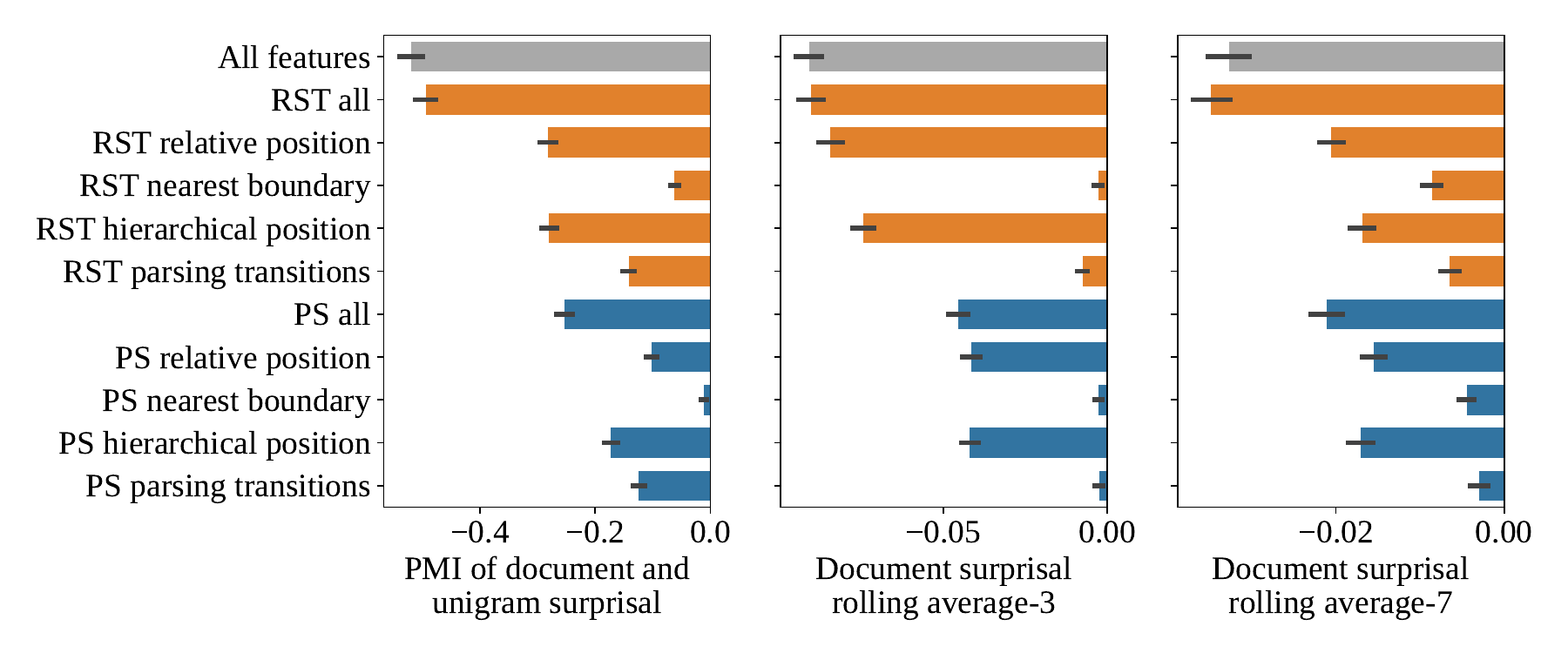}
 \caption{English}
    \label{subfig:comparison_rst_prose_en_extra}
  \end{subfigure}
  \begin{subfigure}[b]{1\textwidth}
  \includegraphics[width=\textwidth]{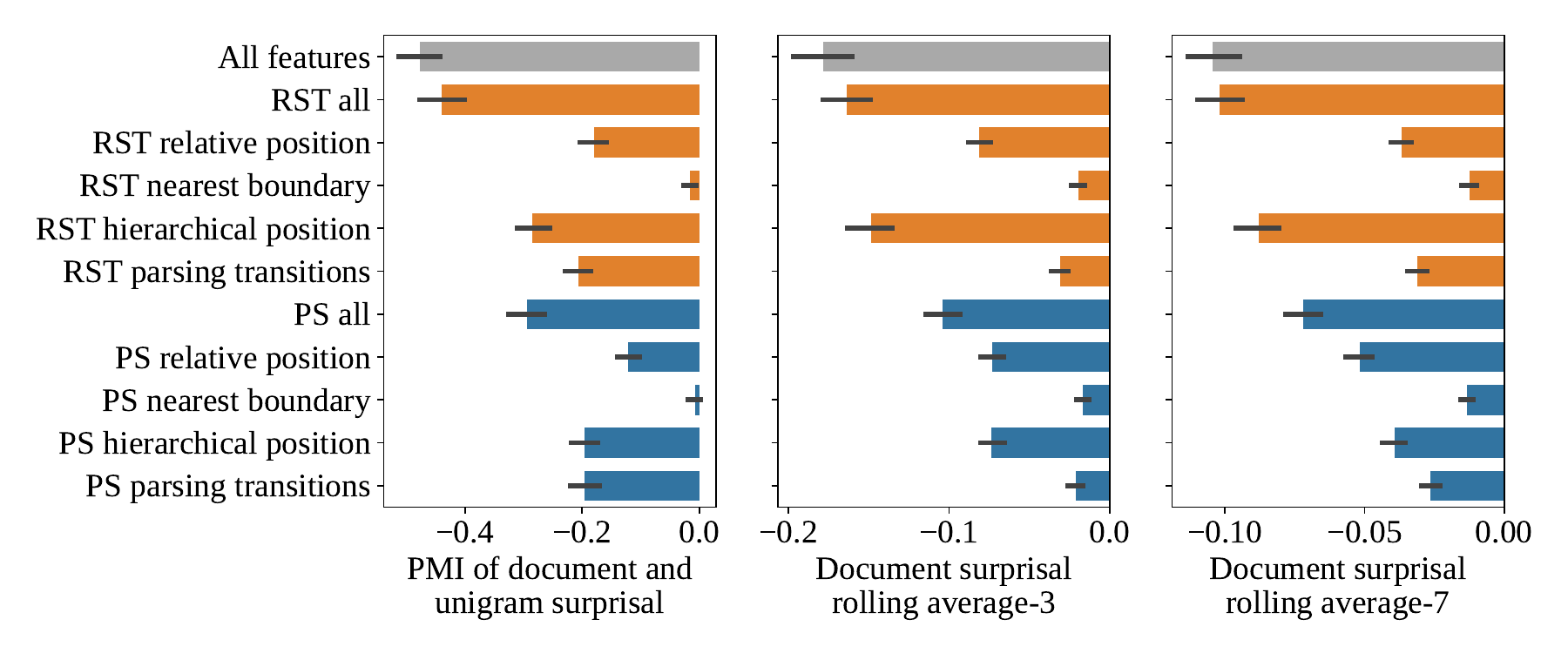}
  \caption{Spanish}
    \label{subfig:comparison_rst_prose_es_extra}
    \end{subfigure}
  \caption{\deltamse\ of RST and Prose Structure (PS) across the remaining dependent variables.}
  \label{fig:comparison_rst_prose_extra}
\end{figure*}

In \Cref{fig:comparison_rst_prose_es}, we show the Spanish results corresponding to the English ones in \Cref{fig:comparison_rst_prose_en}. We also provide the results on the remaining dependent variables in \Cref{subfig:comparison_rst_prose_en_extra} for English and \Cref{subfig:comparison_rst_prose_es_extra} for Spanish. 

\end{document}